\documentclass[10pt,twocolumn,letterpaper]{article}

\usepackage[pagenumbers]{cvpr} 
\usepackage{dsfont}
\usepackage{multirow}
\usepackage{cuted}
\usepackage{caption}
\usepackage{bbding}
\usepackage{makecell}

\usepackage[most]{tcolorbox}
\usepackage{xcolor}
\usepackage{fvextra} 
\usepackage{inconsolata} 

\definecolor{promptbg}{RGB}{245,245,245} 
\definecolor{prompttitlebg}{RGB}{70,70,70} 
\definecolor{prompttitlefg}{RGB}{255,255,255} 

\newtcolorbox{prompttitle}{
    colback=prompttitlebg,
    coltext=prompttitlefg,
    left=6pt, right=6pt, top=6pt, bottom=6pt,
    boxrule=0pt,
    enhanced,
    sharp corners,
}

\newtcolorbox{promptbox}{
    breakable,
    enhanced,
    colback=promptbg,
    colframe=black,
    boxrule=0.5pt,
    sharp corners,
    left=10pt,
    right=10pt,
    top=8pt,
    bottom=8pt,
    fontupper=\ttfamily\small,
}

\DefineVerbatimEnvironment{PromptVerb}{Verbatim}{
    breaklines=true,
    breakanywhere=true,
    breaksymbol={},       
    fontsize=\small,
}

\newtcolorbox{promptblock}[1]{%
    breakable,
    enhanced,
    colback=promptbg,            
    colframe=black,              
    colbacktitle=prompttitlebg,  
    coltitle=prompttitlefg,      
    title=#1,                    
    fonttitle=\bfseries\normalsize,   
    sharp corners,
    boxrule=0.5pt,
    left=10pt,
    right=10pt,
    top=8pt,
    bottom=8pt,
    fontupper=\ttfamily\small,   
}

\def\maketitlesupplementaryoc
   {
   \newpage
       \onecolumn 
       {
        \centering
        \Large
        \textbf{\thetitle}\\
        \vspace{0.5em}Supplementary Material \\
        \vspace{1.0em}
        }
   }

\definecolor{cvprblue}{rgb}{0.21,0.49,0.74}
\usepackage[pagebackref,breaklinks,colorlinks,allcolors=cvprblue]{hyperref}

\usepackage{xcolor}
\usepackage[table]{xcolor}
\definecolor{MyRed}{HTML}{FF0000}
\definecolor{MyBlue}{HTML}{2972F4}

\newcommand{\mred}[1]{{\color{MyRed}#1}}
\newcommand{\mblue}[1]{{\color{MyBlue}#1}}

\def\paperID{XXXX} 
\def\confName{CVPR}
\def\confYear{2026}

\title{Deep But Reliable: Advancing Multi-turn Reasoning for Thinking with Images}

\author{Wenhao Yang\textsuperscript{\rm 1,2,3,}\thanks{Equal Contribution.} \thanks{Work done during an internship at AI Business, Alibaba Group.} , Yu Xia\textsuperscript{\rm 3,}\footnotemark[1] , Jinlong Huang\textsuperscript{\rm 3,4,}\footnotemark[2] , Shiyin Lu\textsuperscript{\rm 3}, Qing-Guo Chen\textsuperscript{\rm 3},  \\ 
Zhao Xu\textsuperscript{\rm 3}, Weihua Luo\textsuperscript{\rm 3}, Kaifu Zhang\textsuperscript{\rm 3}, Yuanyu Wan\textsuperscript{\rm 5} and Lijun Zhang\textsuperscript{\rm 1,2,}\thanks{Corresponding author, email: zhanglj@lamda.nju.edu.cn.} \\
\textsuperscript{\rm 1}National Key Laboratory for Novel Software Technology, Nanjing University, China \\ 
\textsuperscript{\rm 2}School of Artificial Intelligence, Nanjing University, China \\
\textsuperscript{\rm 3}AI Business, Alibaba Group \\
\textsuperscript{\rm 4}School of Automation and Intelligent Sensing, Shanghai Jiao Tong University, China \\
\textsuperscript{\rm 5}School of Software Technology, Zhejiang University, China 
}

\begin{document}
\maketitle

\begin{strip}
\begin{minipage}{\textwidth}
\vspace{-16mm}
\centering
\includegraphics[width=0.98\linewidth]{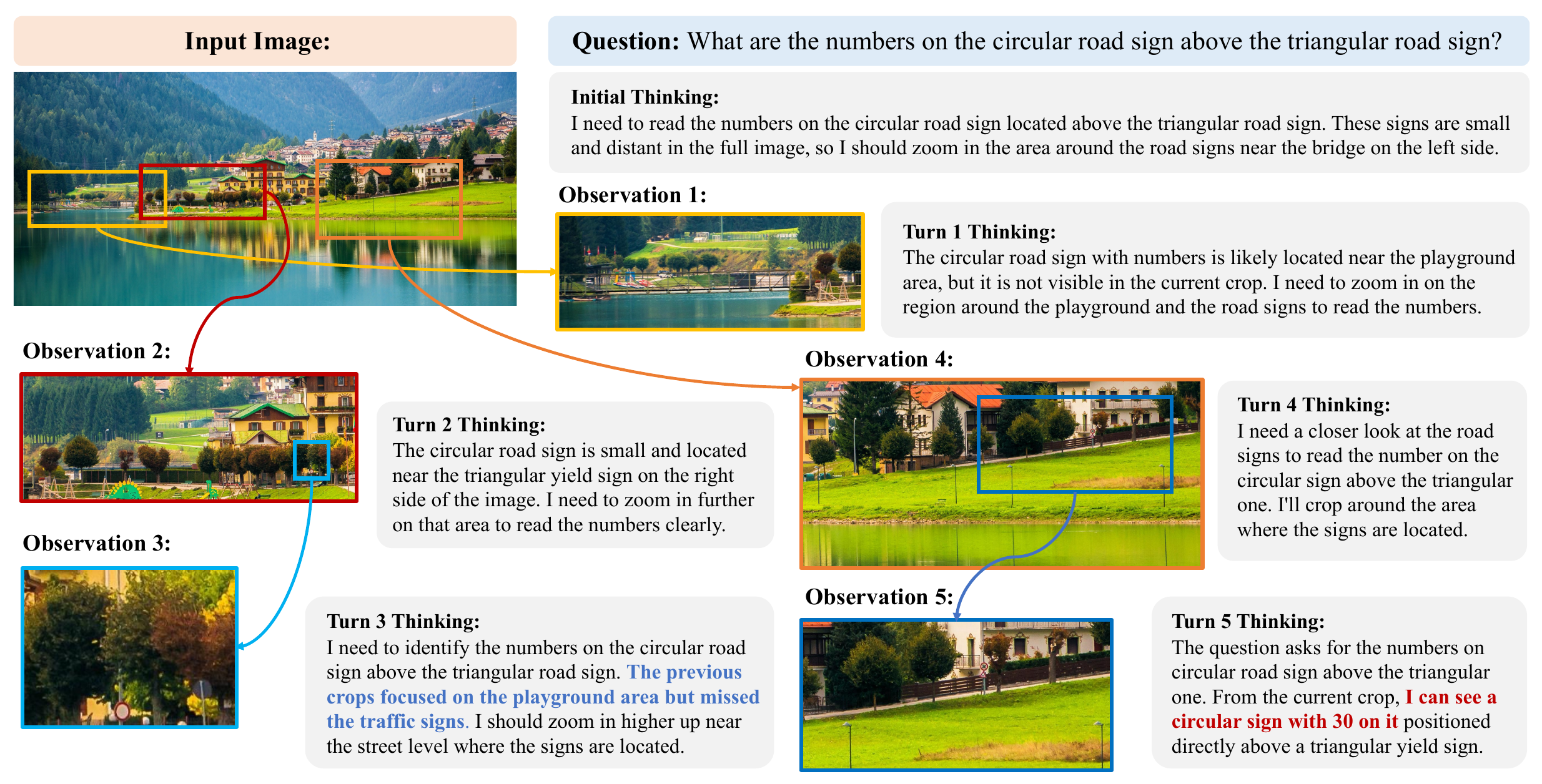}
\captionof{figure}{Illustration of DRIM performing multi-turn reasoning to tackle a visual search task. Our model thinks with images in its MCoT, invoking the zoom-in tool to crop the image and analyze it more thoroughly. In addition, DRIM can reflect and self-correct during the reasoning process (highlighted in \mblue{blue}), thereby localizing the correct region and producing the final answer (highlighted in \mred{red}). }
\label{fig:teaser}
\end{minipage}
\end{strip}

\begin{abstract}
Recent advances in large Vision-Language Models (VLMs) have exhibited strong reasoning capabilities on complex visual tasks by thinking with images in their Chain-of-Thought (CoT), which is achieved by actively invoking tools to analyze visual inputs rather than merely perceiving them. However, existing models often struggle to reflect on and correct themselves when attempting incorrect reasoning trajectories. To address this limitation, we propose DRIM, a model that enables \textbf{D}eep but \textbf{R}eliable multi-turn reasoning when thinking with \textbf{I}mages in its \textbf{M}ultimodal CoT. Our pipeline comprises three stages: data construction, cold-start SFT and RL. Based on a high-resolution image dataset, we construct high-difficulty and verifiable visual question–answer pairs, where solving each task requires multi-turn tool calls to reach the correct answer. In the SFT stage, we collect tool trajectories as cold-start data, guiding a multi-turn reasoning pattern. In the RL stage, we introduce  redundancy-penalized policy optimization, which incentivizes the model to develop a self-reflective reasoning pattern. The basic idea is to impose judgment on reasoning trajectories, and penalize those that produce incorrect answers without sufficient multi-scale exploration. Extensive experiments demonstrate that DRIM achieves superior performance on visual understanding benchmarks. 
\end{abstract}    
\section{Introduction}
\label{sec:intro}

The field of Large Vision-Language Models (VLMs) has witnessed rapid advancements, with numerous open-source models demonstrating remarkable capabilities~\cite{llava-next,Qwen25-VL,Internvl3,GLM-41V,Keye-VL,Ovis25}. To strengthen the reasoning ability of VLMs on complex multimodal inputs, many efforts have introduced a long internal Chain-of-Thought (CoT)~\cite{NeurIPS:2022:Wei,NeurIPS:2022:Kojima}, training models to think longer before answering~\cite{arXiv:2023:Zhang,ACL:2025:He,arXiv:2025:Shen,Mathvista}. Despite these promising results, most existing VLMs remain confined to a text-only reasoning pattern, failing to fully exploit visual information. In particular, these models can only ``see'' images by treating them as static inputs within the CoT, resulting in reasoning processes that are heavily dominated by the language modality. Therefore, researchers have sought to integrate visual information into CoT, extending the traditional text-only reasoning pattern~\cite{NeurIPS:2024:Hu,arXiv:2025:DeepEyes}. They achieve this by transforming original images with tools, such as performing crop, zoom-in or other image manipulation operations, and incorporate the modified images into CoT to enhance visual reasoning. 

Recently, the release of OpenAI-o3 and o4-mini~\cite{openaio3} has marked a milestone in the development of Visual Reasoning Models (VRMs). The o3 model further advances the reasoning capabilities by incorporating visual information as a dynamic cognitive workspace into its CoT reasoning. It can autonomously perform multi-turn tool invocation, enabling more accurate and thorough visual analysis than conventional VRMs. This emerging paradigm, ``\textbf{Thinking with Images}'', has significantly advanced multimodal understanding, revealing the potential for VRMs to exhibit a more holistic and human-like form of cognition~\cite{LARKIN198765}. Given these merits, a pivotal question arises: \emph{How can we incentivize VLMs to ``think with images''?}

\begin{figure}[t]
  \centering
  \begin{minipage}[t]{0.96\linewidth}
    \centering
    \includegraphics[width=\linewidth]{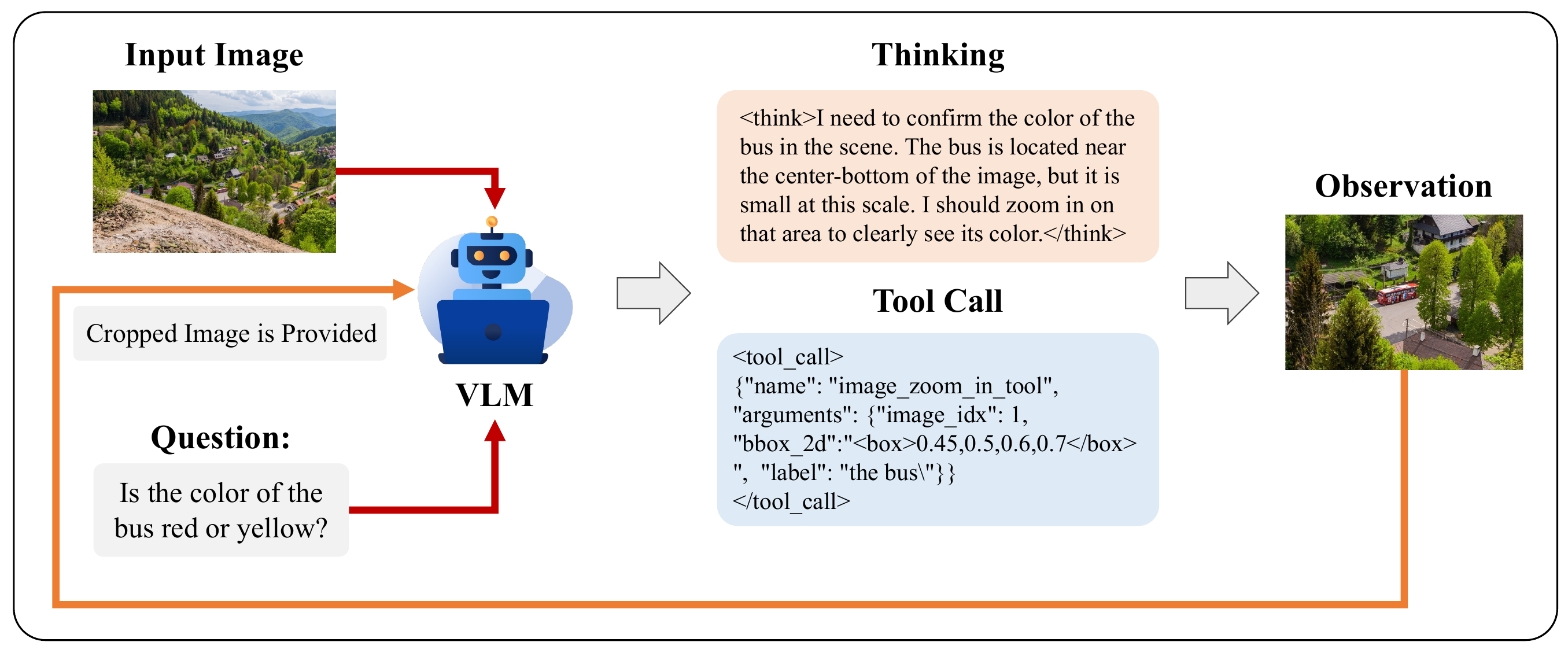}
    \subcaption{Overview of Agentic Pipeline}
    \label{fig:agentic}
  \end{minipage}
  \begin{minipage}[t]{0.96\linewidth}
    \centering
    \includegraphics[width=\linewidth]{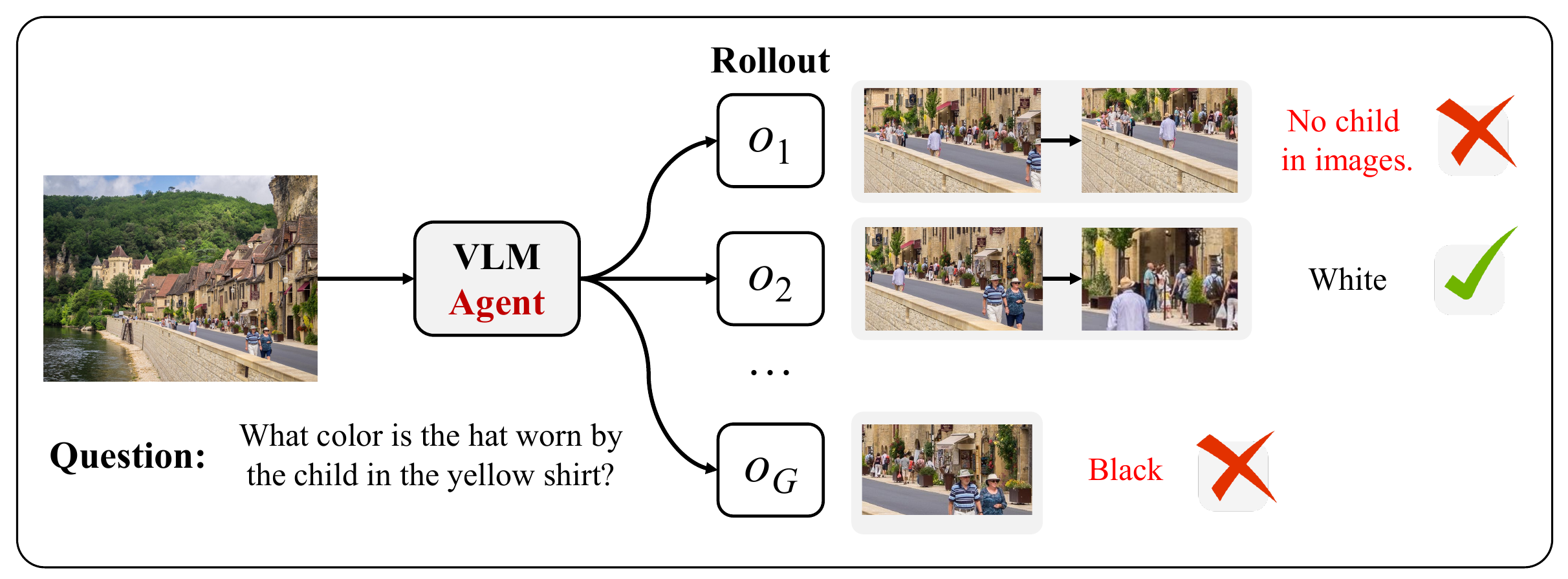}
    \subcaption{Reward Signal in RL training}
    \label{fig:intro2}
  \end{minipage}
  \caption{
    The illustration of the agentic pipeline and reward signal in RL training under the paradigm of ``Thinking with Images''.
  }
\end{figure}

Motivated by the remarkable progress of reinforcement learning (RL) in language reasoning models, such as OpenAI-o1~\cite{openaio1} and DeepSeek-R1~\cite{DeepSeek-r1}, a natural solution is to adopt RL for enabling multi-turn multimodal reasoning. Pioneering open-source work, DeepEyes~\cite{arXiv:2025:DeepEyes}, proposes an end-to-end RL training recipe. Their method follows an agentic pipeline, as shown in \Cref{fig:agentic}. Given an image and a question, the VLM acts as an agent that iteratively produces a \emph{thinking} text and a \emph{tool} call function. The function triggers a tool to operate on the image, resulting in a new image called an \emph{observation}. This observation, together with all historical information, is fed back into the VLM in the next turn, continuing until a final answer is generated. In the RL stage, DeepEyes treats the correctness of the final answer as the reward signal, and uses policy optimization to enhance the reasoning trajectory accordingly. 

However, DeepEyes, along with most recent recipes~\cite{arXiv:2025:Thyme,arXiv:2025:Minio3}, still falls short of achieving \emph{reliable} multi-turn reasoning. Specifically, the model usually attempts different visual reasoning trajectories for a problem, some of which lead to incorrect answers. Meanwhile, it struggles to engage in self-reflection and correction during the reasoning process. This unreliability largely stems from the RL training scheme, where the reward signal only encourages rollout trajectories that yield correct answers without any mechanism to evaluate the reasoning process itself. As a result, the model tends to guess an answer with fewer reasoning turns rather than developing multi-turn and self-reflective reasoning patterns, as is shown in~\Cref{fig:intro2}.  

To address this, we propose \textbf{DRIM}, a model that enables \textbf{D}eep but \textbf{R}eliable multi-turn reasoning pattern when thinking with \textbf{I}mages in its \textbf{M}ultimodal CoT. 
Our overall pipeline for implementing DRIM consists of three stages: high-quality data construction, cold-start Supervised Fine-Tuning (SFT), and end-to-end reinforcement learning (RL). Based on a high-resolution image dataset, we design an automated scheme to generate \emph{high-difficulty} and \emph{verifiable} visual reasoning dataset, i.e., multimodal Question–Answer (QA) pairs. Compared with existing training datasets~\cite{arXiv:2025:DeepEyes,arXiv:2025:Minio3}, our proposed dataset requires the model to perform multiple tool calls to arrive at the correct answer, thereby effectively incentivizing ``thinking with images''. During the SFT stage, we collect tool-call trajectories associated with the QA pairs as cold-start data, guiding the model to acquire multi-turn reasoning pattern and tool-calling abilities. During the RL stage, the model performs multiple rollouts of exploration and reasoning process, with its policy updated according to the reward signals. To promote the self-reflective reasoning patterns, we introduce  redundancy-penalized policy optimization. The basic idea is to impose additional judgment on reasoning trajectories, and penalize those that produce incorrect answers without engaging in sufficient multi-scale exploration. For visual understanding tasks, DRIM demonstrates superior reasoning capabilities, which is demonstrated by extensive experiments.  

\Cref{fig:teaser} demonstrates the remarkable performance of DRIM in complex visual reasoning scenarios. When faced with cluttered images containing redundant information, most VRMs can only perceive the raw image and rely on single-step, text-only reasoning grounded in static visual features, which often fails to locate the correct target. In contrast, DRIM can iteratively zoom in and crop regions of interest, gradually refining its focus and accurately identifying the target through multi-turn reasoning.

Our contributions can be summarized as follows:
\begin{itemize}
    \item We construct a new multimodal reasoning dataset that follows two key principles, \emph{high-difficulty} and \emph{verifiability}, encouraging models to invoke tools for visual reasoning.
    \item We refine the training scheme for promoting ``thinking with images'', consisting of a cold-start SFT stage and an end-to-end RL stage.
    \item We introduce redundancy-penalized policy optimization, which incentivizes the model to develop self-reflective reasoning patterns through broader-scale exploration.
\end{itemize}

\begin{figure*}[t]
\begin{center}
\centerline{\includegraphics[width=.96\linewidth]{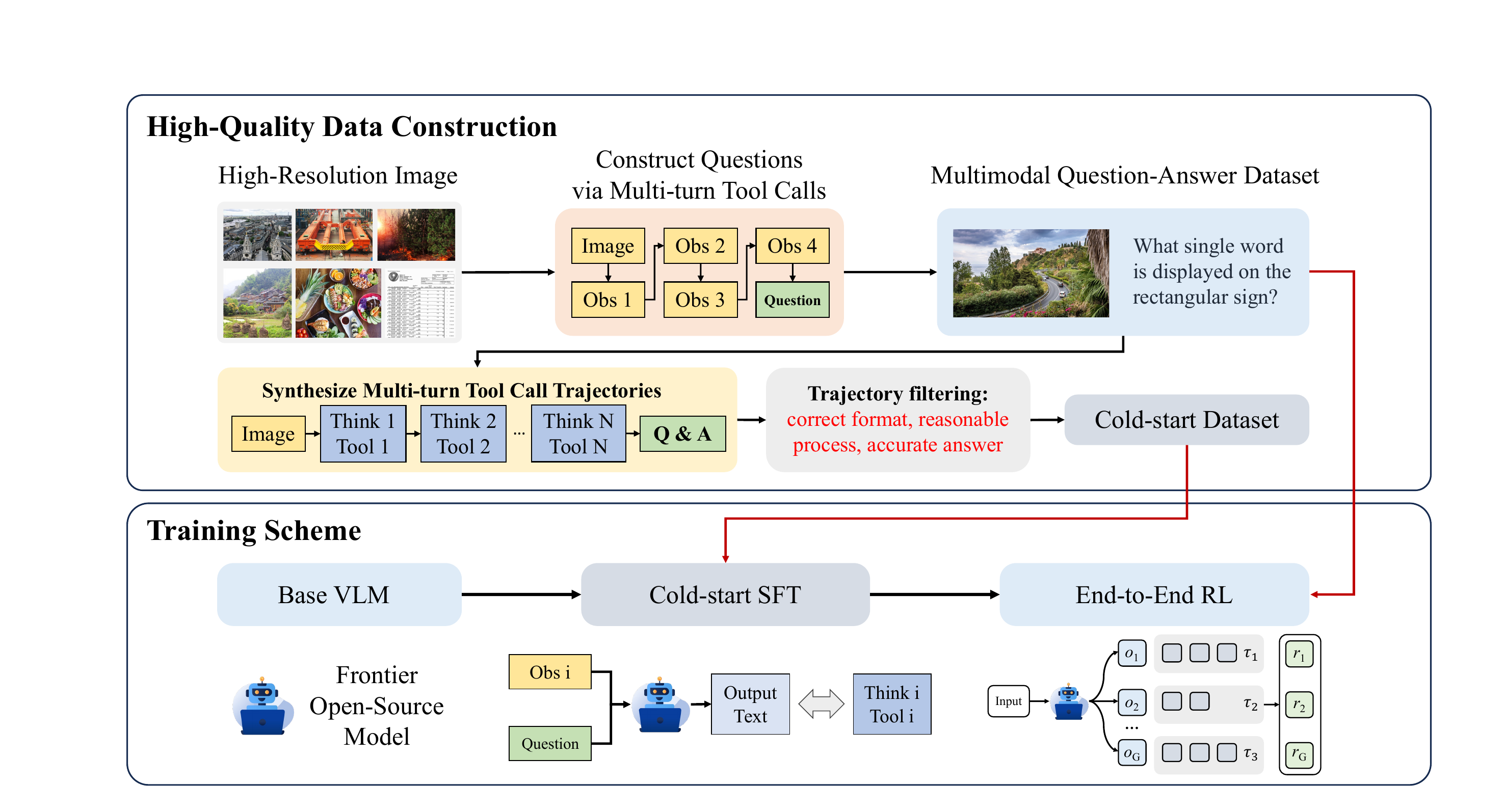}}
\caption{Overview of the overall pipeline for implementing DRIM. Our pipeline comprises three stages: data construction, cold-start SFT and RL. First, we construct a new multimodal dataset, and synthesize multi-turn tool call trajectories to serve as cold-start data. Second, the synthesized trajectories are used to SFT the model, enabling it to acquire tool-use abilities and multi-turn reasoning. Third, we design reward signals that encourage the model to autonomously explore and optimize its reasoning trajectories during RL training.}
\label{fig:pipeline}
\end{center}
\end{figure*}

\section{Related Work}
\paragraph{Large Vision-Language Models.} Large VLMs, capable of jointly perceiving visual and linguistic modalities, have become a central research focus in recent years. Early works, such as CLIP~\cite{NeurIPS:2021:CLIP,ICCV:2023:Siglip} and BLIP~\cite{ICML:2022:BLIP,ICML:2023:BLIP2}, trained vision–language representations from scratch using large-scale image–text pairs. However, these approaches lack strong contextual understanding and are unable to process multiple images. To address these limitations, subsequent representative models including Flamingo~\cite{NeurIPS:2022:Flamingo} and LLaVA~\cite{NeurIPS:2023:LLaVA} build on  powerful pre-trained large language models (LLMs) as backbones, aligning visual and textual information by connecting a pre-trained ViT~\cite{ViT} to the LLM through a simple projector like MLP. Driven by the rapid advancement of LLMs, a growing family of open-source VLMs has emerged, including LLaVA~\cite{LLaVA15,llava-next,Llava-onevision}, InternVL~\cite{internvl,internvl15,internvl25,Internvl3,internvl35}, Qwen-VL~\cite{qwenvl,qwen2vl,Qwen25-VL}, Ovis~\cite{ovis,Ovis25}, SEED-VL~\cite{seed15vl}, and GLM-VL~\cite{GLM-41V}. These models demonstrate strong capabilities across a wide range of visual tasks, such as visual question answering, image grounding, and fine-grained image understanding. 

\paragraph{Think with Images.} Most existing VRMs have achieved some success by introducing CoT reasoning into VLMs~\cite{Mathvista,arXiv:2023:Zhang}, yet they remain confined to text-dominant reasoning paradigms. To further advance multimodal reasoning, the ``thinking with images'' paradigm has gained increasing attention. Instead of treating visual information as a static input, this paradigm incorporates visual signals into the CoT as dynamic components of the reasoning workflow. Early studies adopt external tools to generate new images during reasoning~\cite{NeurIPS:2024:Hu,arXiv:2025:Huang}, thereby forming a multimodal CoT that enhances the model’s reasoning capability. Subsequent tool-driven approaches for incentivizing ``thinking with images'' can be categorized into three main families: prompt-based~\cite{EMNLP:2025:zoomeye,arXiv:2023:Yang,arXiv:2025:Cheng}, SFT-based~\cite{ECCV:2024:Liu,arXiv:2025:Bai,corr:2024:Shao,arXiv:2025:Zhang,CVPR:2024:V*}, and RL-based methods~\cite{arXiv:2025:Liu,arXiv:2025:Zhu,arXiv:2025:Zhang:B,arXiv:2025:Su,arXiv:2025:Minio3}. Among them, DeepEyes~\cite{arXiv:2025:DeepEyes} introduces an end-to-end RL framework. Unlike early approaches that rely on predefined workflows, DeepEyes leverages the model’s native grounding ability to support MCoT, offering improved flexibility and scalability. Undoubtedly, the ``thinking with images'' paradigm is unlocking the full multimodal potential of VLMs. 

\section{Methodology}
Given an image input and a question query, the model typically performs multiple reasoning attempts, some of which will lead to incorrect results. Therefore, to achieve stable reasoning, we aim to endow the model with a deep but reliable reasoning pattern. Here, ``Deep'' refers to the model’s ability to conduct multi-turn reasoning, while ``Reliable'' denotes its capacity for self-reflection and self-correction. However, most existing models for reproducing ``thinking with images'' lack such a reasoning pattern. When faced with challenging problems, these models tend to produce vague answers within few turns, which can be regarded as guessing an answer to obtain potential rewards. To address this limitation, this work focuses on incentivizing the model to acquire a deep but reliable reasoning pattern. 

To this end, we propose our DRIM, whose base model is built upon the open-source VLM Ovis2.5-9B~\cite{Ovis25}. The overview of our pipeline for implementing DRIM is introduced in Section~\ref{sec:pipeline}. Instead of using currently available open-source training datasets in which the questions can be solved without invoking any external tools, we construct an entirely new collection of high-quality and diverse visual question–answer data as our cold-start and RL training dataset, which is detailed in Section~\ref{sec:data}. For the training scheme, we first use multi-turn trajectory data to cold-start the base model, and then employ end-to-end RL to further enhance the  reasoning capability of our model, which are detailed in Section~\ref{sec:trainer}.

\subsection{Overview of Pipeline}\label{sec:pipeline}
Our overall pipeline for implementing DRIM consists of three stages, including high-quality data construction, cold-start SFT, and end-to-end RL training. \Cref{fig:pipeline} provides an overview of the pipeline, and illustrates the relationships among different stages to facilitate understanding. 

The overall pipeline starts with high-quality data construction, which serves as a crucial component for incentivizing the model to think with images in its MCoT. Based on a high-resolution image dataset, we construct visual understanding questions through multi-turn tool calls, which subsequently guide the model to learn step-by-step reasoning on complex problems during the training stage. We design an automated procedure to generate visual question and answer (QA) pairs, thereby building a multimodal QA dataset. Subsequently, using the constructed dataset, we employ frontier VLMs such as o4-mini to synthesize multi-turn tool call trajectories, followed by manual trajectory filtering. Finally, the synthesized trajectories are used as cold-start data, while the multimodal QA dataset serves as the training data for RL.

For the training scheme, we first require a strong open-source VLM as the base model so that its native grounding and reasoning capabilities can be leveraged to incentivize thinking with images within its MCoT. In this work, we adopt Ovis2.5-9B~\cite{Ovis25} as the base model. During the SFT stage, we perform a cold-start initialization that enables the model to acquire fundamental tool-use abilities and multi-turn reasoning pattern. With the synthesized multi-turn trajectories, SFT ensures that the model can stably invoke visual tools and is capable of using code-based tools to accomplish image manipulation. Subsequently, in the RL stage, the model no longer relies on the synthesized trajectories. Instead, we design reward signals that encourage the model to autonomously explore and optimize its reasoning trajectories. RL training is crucial for enabling the model to perform multi-turn tool calls and acquire visual information in a self-directed manner, thereby realizing an OpenAI-o3-style multimodal reasoning pattern. 

\subsection{Data Construction}\label{sec:data}
\begin{figure}[t]
    \centering
    \includegraphics[width=0.98\linewidth]{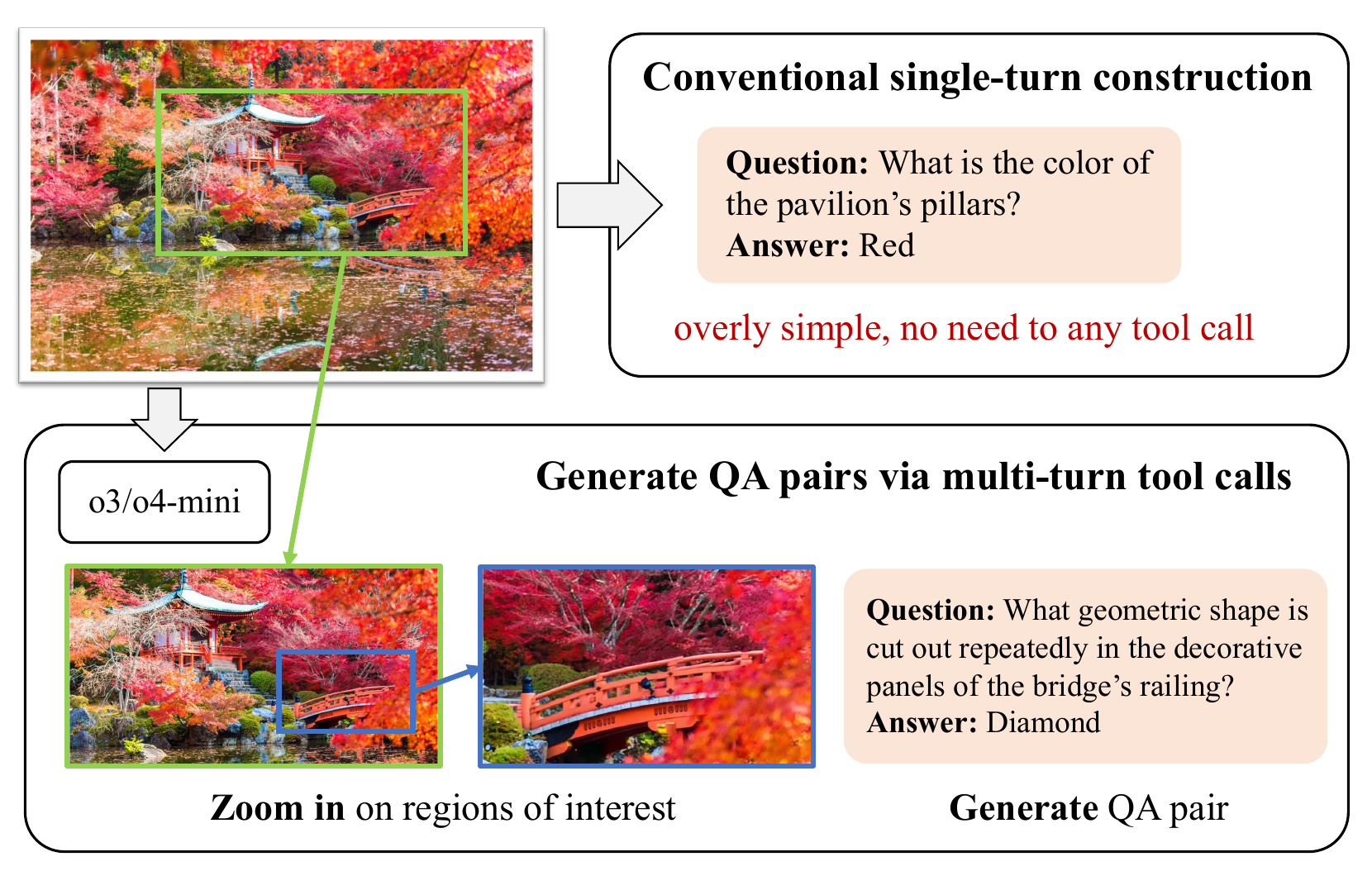}
    \caption{Our automated scheme for data construction. In our scheme, the frontier VLMs select and zoom into the regions of interest, and then generate QA pairs on the specific region.}
    \label{fig:data}
\end{figure}

To effectively guide the model’s visual reasoning capability, we design an automated program to generate a collection of \emph{high-difficulty} and \emph{verifiable} multimodal training data. It is well known that constructing datasets for visual reasoning tasks is both highly challenging and crucial, as training data that can incentivize the model to think with images is required to satisfy the following two principles:
\begin{itemize}
    \item \textbf{High-Difficulty}: Visual reasoning tasks must possess sufficient complexity, such that the model cannot easily arrive at the correct answer without the assistance of visual tools. This encourages the model to engage in multi-turn visual reasoning.
    \item \textbf{Verifiability}: The generated question–answer pairs must be verifiable and trustworthy, guiding the model to produce reliable reasoning trajectories rather than degenerate optimization behaviors.
\end{itemize}
Existing open-source training datasets generally struggle to meet these two principles. Specifically, most training samples can be solved without invoking external tools, making it difficult to guide the model to perform multi-turn visual reasoning. To address this, we design a data construction scheme based on high-resolution images, where frontier visual reasoning models such as o4-mini are employed to automatically generate questions. Our scheme is demonstrated in \Cref{fig:data}, where the o3/o4-mini model iteratively selects regions of interest in the original image, progressively zooms in on them over multiple turns, and then generates corresponding question–answer pairs conditioned on the final zoomed-in view. This procedure not only ensures sufficient task difficulty, but also preserves answer verifiability through localized visual details. 

\subsection{Training Scheme}\label{sec:trainer} 

\paragraph{Cold-start SFT.} The importance of cold-start SFT lies in its ability to equip the model with fundamental tool-use abilities and multi-turn reasoning pattern. Notably, previous work, such as the representative DeepEyes~\cite{arXiv:2025:DeepEyes}, directly employed RL without cold-start, where tool call and instruction-following capabilities were solely incentivized by reward signals. However, we observe that models without cold-start  tend to produce concise and vague answers with very few reasoning turns, a phenomenon also noted in a concurrent work~\cite{arXiv:2025:Minio3}. For complex visual reasoning tasks, relying entirely on reward signals in RL is insufficient to activate multi-turn reasoning trajectories that lead to correct answers. Therefore, we utilize synthesized tool call trajectories as cold-start data to train a base model with preliminary thinking with images capability, establishing a solid foundation for subsequent RL training.

\paragraph{Agentic RL.} Following the rollout formulation of DeepEyes~\cite{arXiv:2025:DeepEyes}, we draw inspiration from agentic RL to formulate the ``thinking with images'' reasoning pattern as a Markov Decision Process (MDP) that incorporates environmental feedback. In contrast to traditional RL with text-only CoT, our formulation introduces observation tokens, thereby forming a Multimodal CoT (MCoT). 

At each step $t$ in MCoT, the system state $s_t$ is no longer limited to the model’s own historical outputs but instead encompasses the complete interaction history, including both the model’s generated reasoning content and the environmental feedback. Specifically, the state $s_t$ is defined as:
\begin{equation*}
    s_t = \{(Q,I_0),(X_1,O_1),(X_2,O_2),\cdots,(X_t,O_t)\},
\end{equation*}
where $(Q,I_0)$ denote the original user question and image, $X_i$ is the think and tool call (e.g., crop, zoom-in) generated by the model, and $O_i$ is the image returned by the environment after executing the tool command in $X_i$. Given the complete current state $s_t$, the  objective of the model is to generate the optimal next-turn thinking and tool call $X_{t+1}$. 

\begin{figure}[t]
  \centering
  \begin{minipage}[t]{0.96\linewidth}
    \centering
    \includegraphics[width=\linewidth]{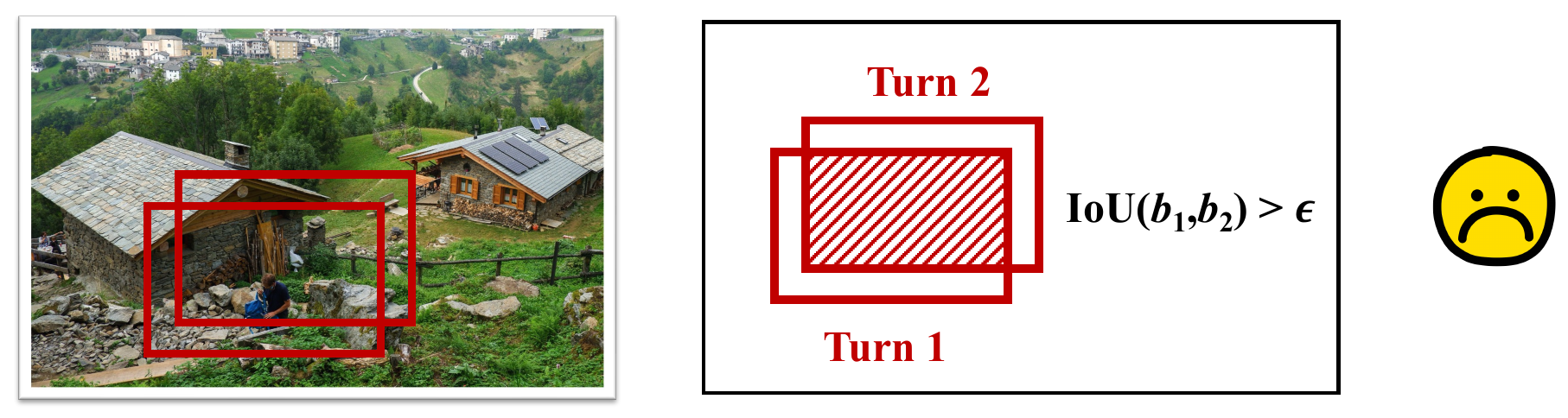}
    \subcaption{Oscillatory Micro-adjustments}
    \label{fig:RLa}
  \end{minipage}
  \begin{minipage}[t]{0.96\linewidth}
    \centering
    \includegraphics[width=\linewidth]{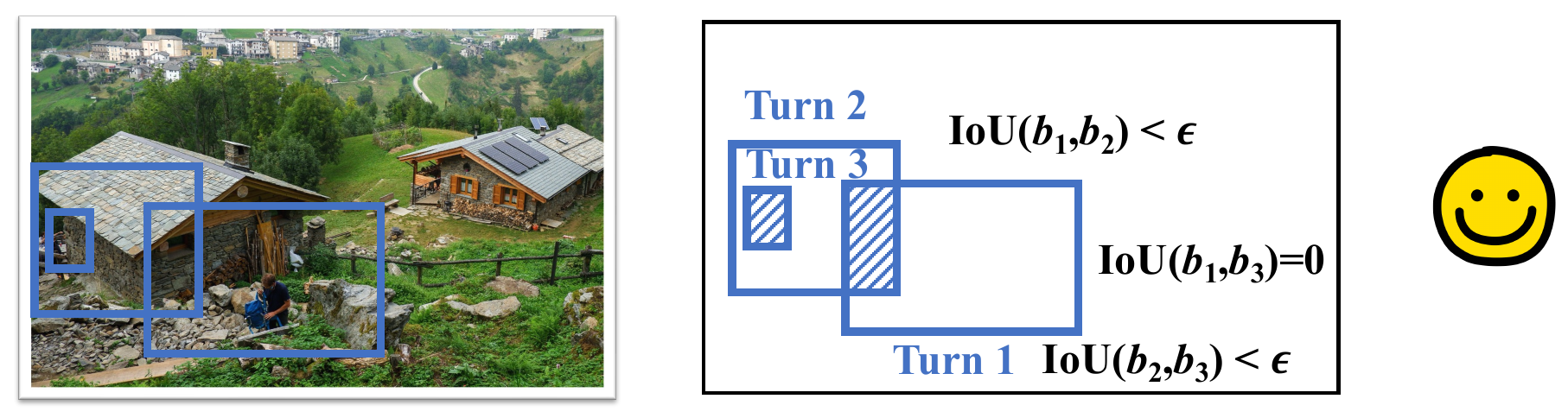}
    \subcaption{Multi-scale Exploration}
    \label{fig:RLb}
  \end{minipage}
  \caption{
    Illustration of multi-turn reasoning trajectories involving zoom-in tool calls. Our objective in optimizing the policy model is to (a) discourage the model from making oscillatory micro-adjustments around the same target, while (b) encouraging to engage into multi-scale exploration.
  }
\end{figure}

\begin{table*}[t]
\caption{\textbf{Main Results on visual understanding datasets}. Best and second best performance are highlighted in  \mred{red} and \mblue{blue}, respectively. $^*$ denotes results reported in the original or other relevant paper. $^{\dagger}$ denotes results reproduced by ourselves through available model weights. Our DRIM achieves the best or comparable performance across three benchmarks.}
\label{tab:main}
\renewcommand{\arraystretch}{1.1}
\setlength{\tabcolsep}{6pt}
\centering
\begin{tabular}{c|cccc|ccc|ccc}
\toprule
\multirow{2}{*}{\textbf{Models}}                    & \multicolumn{4}{c|}{\textbf{VisualProbe}}                       & \multicolumn{3}{c|}{\textbf{V*}} & \multicolumn{3}{c}{\textbf{HR-Bench}}            \\ 
& hard & medium & easy & overall & attribute & relative & overall & 4K & 8K & overall \\
\midrule
GPT-5~\cite{GPT5}    & 31.1   &  22.0  &  57.5 & 33.6 &  72.2 & 77.6 & 74.3 &  74.2  & 72.4 & 73.3    \\ 
Gemini 2.5 Pro~\cite{gemini25} & 29.3   &  36.6  & 53.2 &  39.6 &  87.8 & 72.4 & 81.7 &  68.5  & 60.1  & 64.3 \\ \midrule
LLaVA-OneVision$^*$~\cite{Llava-onevision}   & 13.4 & 12.5 & 36.2 & - & - & - & 70.9 & 61.2 & 54.0 & 57.6     \\
Ovis2.5-9B~\cite{Ovis25} &  12.3 &  26.9 &  51.1  & 31.0  &   81.7 & 78.9  &  80.6 &  72.9 &    67.1  &   70.0         \\ \midrule
SEAL$^*$~\cite{CVPR:2024:V*} &  -   &  -    &  -   & - & 74.8 & 76.3 & 75.4  & -    & - & - \\
DyFo$^*$~\cite{CVPR:2025:Dyfo} &  -   &  - &   - & -  & 80.0 & 82.9 & 81.2  &  -   & - & - \\ 
Thyme$^*$~\cite{arXiv:2025:Thyme} & - & -& -& -& 83.5 & 80.3 & 82.2 & 77.0 & 72.0 & 74.5 \\
DeepEyes$^{\dagger}$~\cite{arXiv:2025:DeepEyes} & 37.8	& 31.3	& \mblue{66.7} &	42.3 & \mblue{90.4} & 88.2  & 89.5	& 75.5	& 71.3 & 73.4  \\ 
DeepEyesv2$^*$~\cite{arXiv:2025:DeepEyesv2} & - & - & - & - & - & - & 81.8 & \mblue{77.9} & \mblue{73.8}  & \mblue{75.9} \\
Mini-o3$^{\dagger}$~\cite{arXiv:2025:Minio3} &   \mred{47.2} & \mblue{45.2} & 64.5 & \mblue{50.9} & \mblue{90.4} & \mblue{92.1} & \mblue{91.1} & 73.5 & 73.0  & 73.3 \\ \midrule
\rowcolor[HTML]{E3E3E3}
\textbf{DRIM} (Ours) &  \mblue{45.3} &	\mred{48.1} &	\mred{69.5} & \mred{53.4} & \mred{91.3}	& \mred{93.4} & \mred{92.2}  & \mred{83.3} &	\mred{82.9}  & \mred{83.1}  \\ 
\rowcolor[HTML]{E3E3E3}
$\Delta$ (\textit{vs} Base Model) & +33.0 & +21.2 & +18.4 & +22.4 & +9.6 & +14.5 & +11.6 & +10.4 & +15.8 & +13.1 \\
\bottomrule
\end{tabular}
\end{table*}

\paragraph{Redundancy-Penalized PO.} To guide the model in exploration and optimization, we need to design an appropriate reward signal during the RL stage. Most existing methods adopt a result-oriented strategy, where the LLM evaluates only the final answer without assessing the intermediate reasoning steps. Such a strategy proves effective for text-only reasoning models, as these models naturally attempt diverse reasoning paths. However, in the context of visual reasoning, the model heavily relies on grounding-based initialization and struggles to perform human-like self-reflection during multi-turn reasoning. Most of failure cases are demonstrated in \Cref{fig:RLa}, where the model often makes oscillatory micro-adjustments around the same target. Ideally, our goal is to encourage the model to conduct multi-scale exploration, as shown in \Cref{fig:RLb}.

To this end, we introduce \emph{redundancy-penalized} policy optimization, where the reward additionally evaluates the quality of multi-turn reasoning trajectories. Our basic idea is to penalize those that produce incorrect answers without engaging in sufficient multi-scale exploration, which motivates the definition of the \emph{redundancy-penalty} term: 
\begin{equation*}
    \Gamma_{rdn}(\tau) = -\frac{\lambda}{\binom{T}{2}} \sum_{t<t^\prime} \max (0,\text{IoU} (b_t,b_{t^\prime})-\epsilon)
\end{equation*}
where $\tau$ denotes the reasoning trajectories, $T$ is the number of tool calls, and $b_t$ is the zoom-in tool box at step $t$ (normalized to the coordinates of the original image and set to None if no tool is used). $\text{IoU}(b_t, b_{t^\prime})$ refers to the \emph{intersection-over-union} between two zoom-in boxes; $\epsilon$ is the tolerance threshold for overlap within the search range, and $\lambda$ is a hyper-parameter. Formally, the final reward is defined as:
\begin{equation*}
    R(\tau) = R_{acc}(\tau) +  \mathds{1} \{ R_{acc}(\tau)=0 \land T>1\}\cdot \Gamma_{rdn}(\tau),
\end{equation*}
where the indicator function activates the redundancy-penalty term only when the final result is incorrect and the reasoning trajectory contains more than one tool call action. 

\paragraph{Implementation.} We implement an RL training framework based on verl~\cite{verl} that supports visual tool invocation. Our framework fully realizes the complete invocation pipeline, including action parsing (model generation of $X_t$), tool execution (environment processing of tool calls), environmental feedback (environment generation of $O_t$), and state updating (integrating $(X_t,O_t)$ into the state $s_t$). 

For the RL algorithm, we adopt Group Relative Policy Optimization (GRPO)~\cite{GRPO}, while other advanced RL methods~\cite{reinforce++,RLOO,dapo} can also be used. For multi-turn trajectories, we apply a loss mask to exclude the loss computation on observation tokens that are not generated by the model~\cite{arXiv:2025:DeepEyes}. 

\begin{figure*}
    \centering
    \begin{minipage}[t]{0.96\linewidth}
    \centering
    \includegraphics[width=\linewidth]{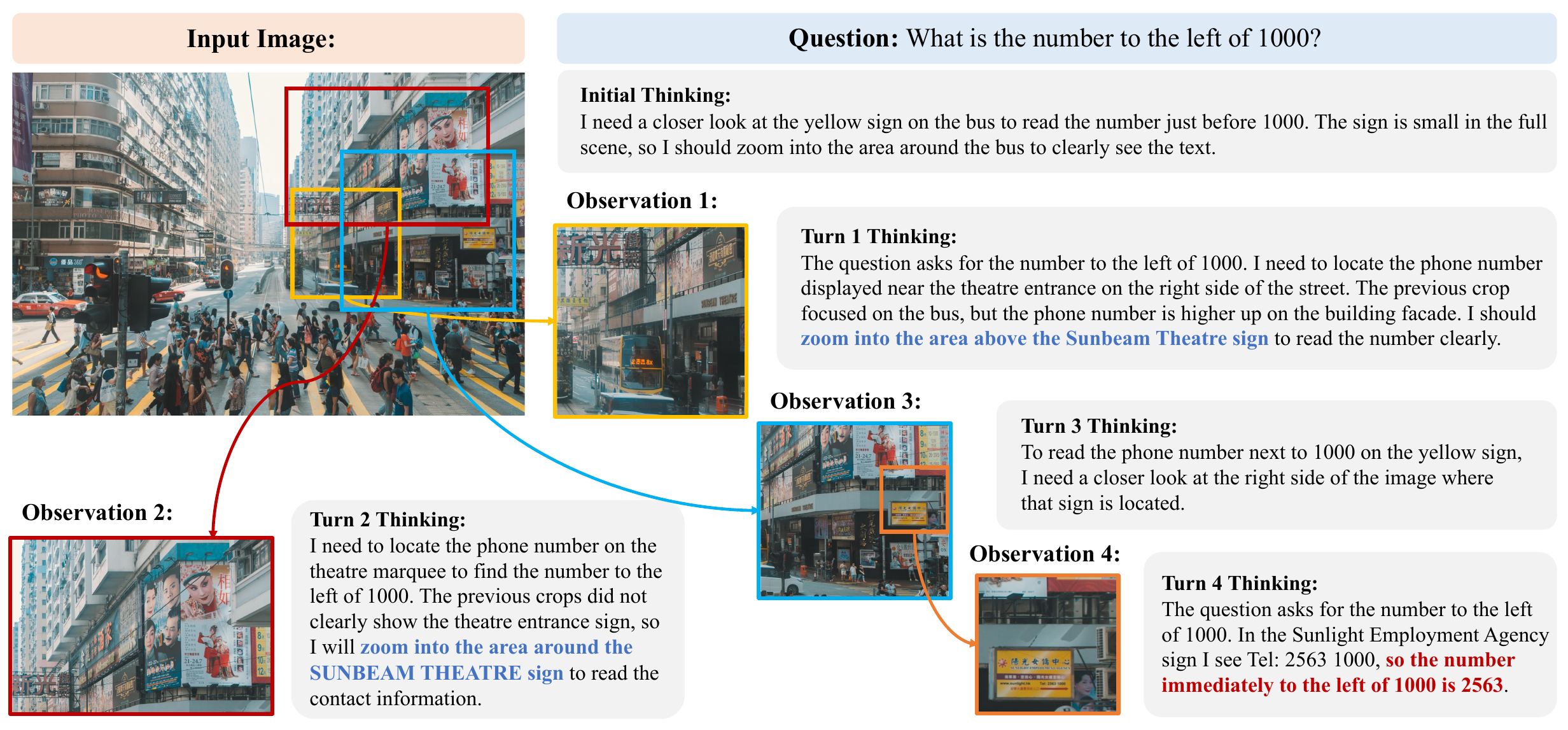}
    \subcaption{Reasoning Trajectories from RL with Redundancy-Penalty}
    \label{fig:Ref1}
    \end{minipage}
    \begin{minipage}[t]{0.96\linewidth}
    \centering
    \begin{minipage}[t]{0.48\linewidth}
            \centering
            \includegraphics[width=\linewidth]{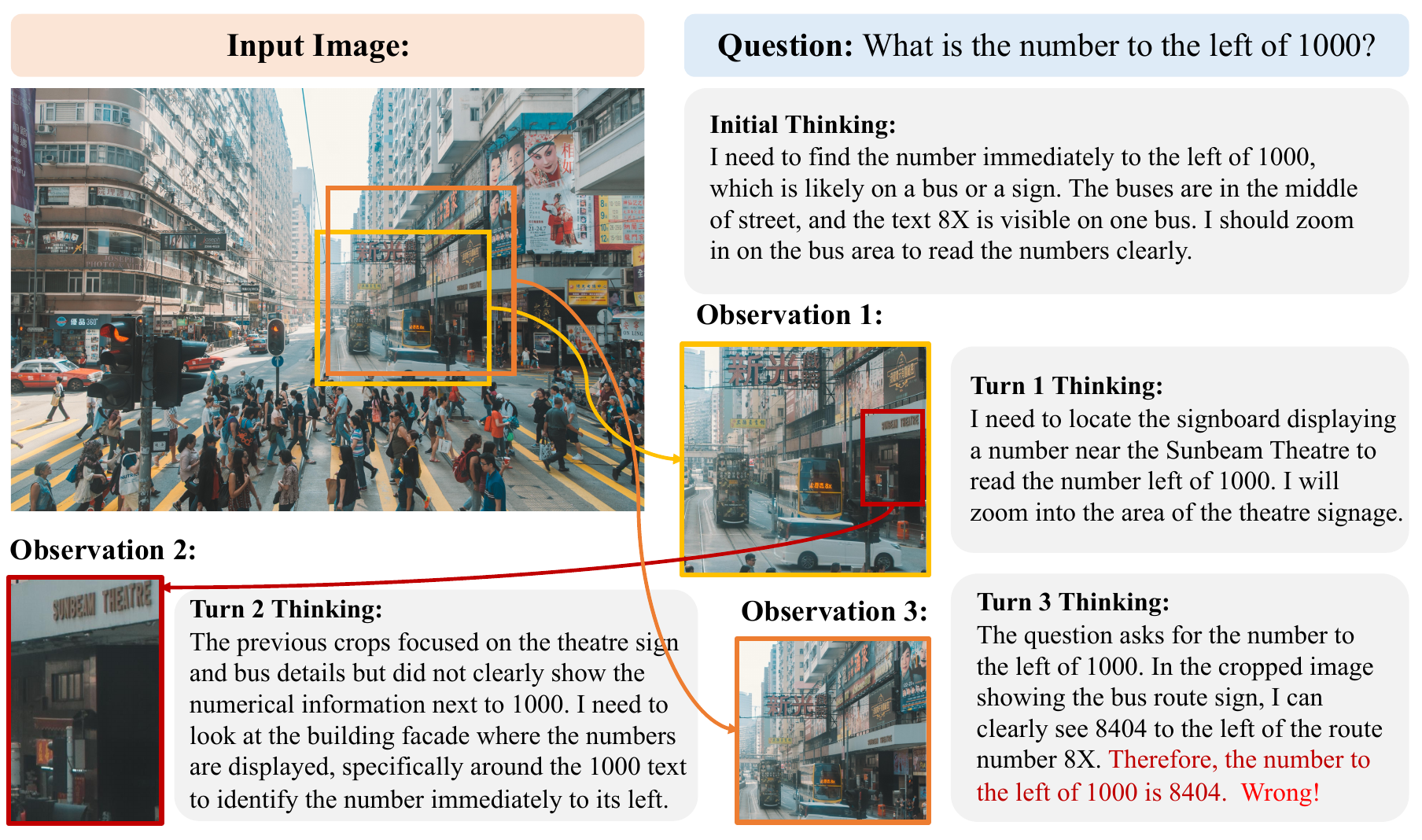}
            \subcaption{Reasoning Trajectories from RL without Redundancy-Penalty}
            \label{fig:Ref21}
        \end{minipage}
        \hfill
        \begin{minipage}[t]{0.48\linewidth}
            \centering
            \includegraphics[width=\linewidth]{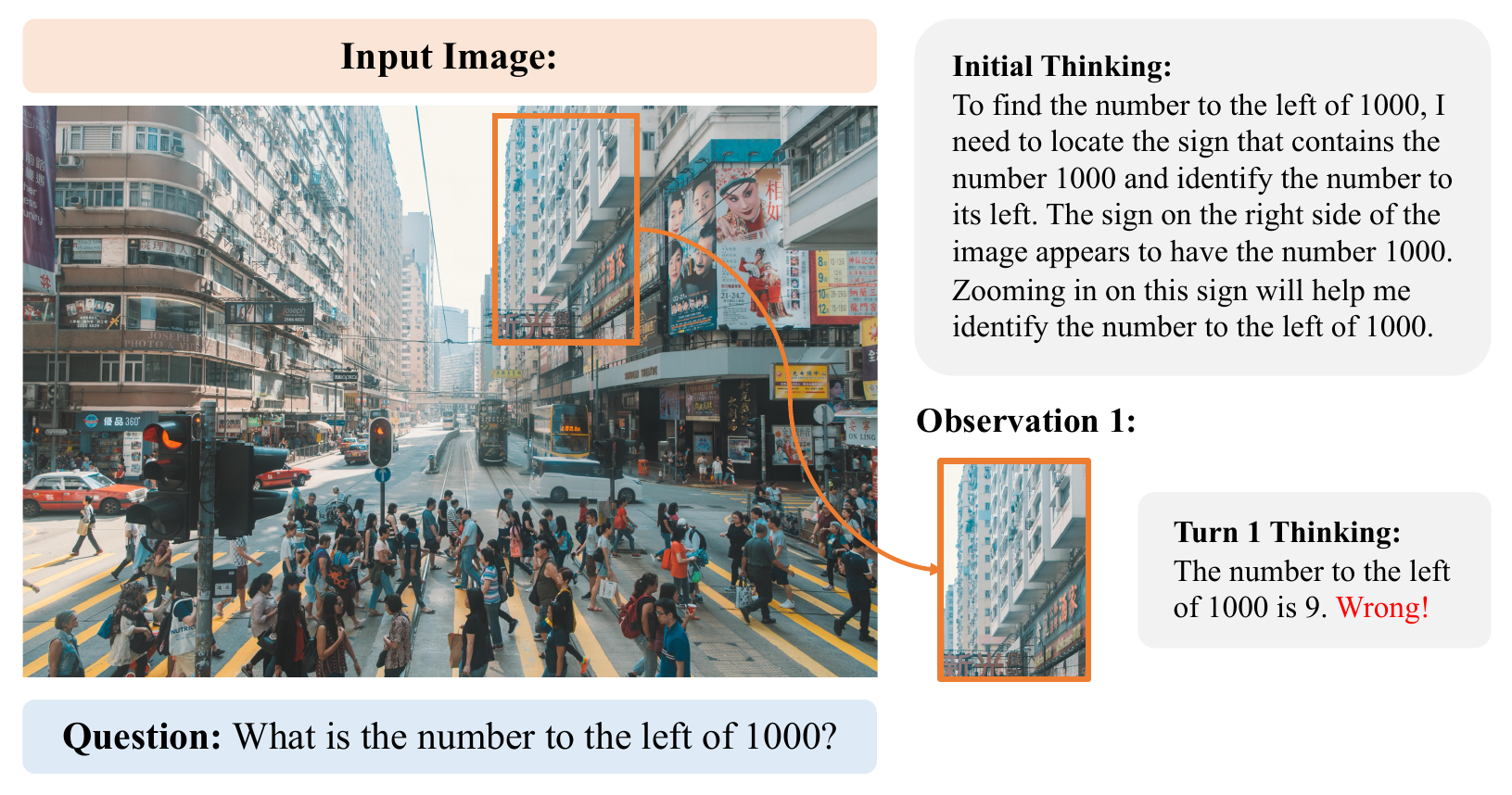}
            \subcaption{Reasoning Trajectories from Other Baseline}
            \label{fig:Ref22}
        \end{minipage}
    \end{minipage}
    \caption{Illustration of different methods performing multi-turn reasoning to solve a visual search task. Compared with other approaches, our method with redundancy-penalty can engage in self-reflection during reasoning (highlighted in \mblue{blue}), thereby enabling broader multi-scale exploration. As shown, distinct reasoning trajectories lead to different final answers (highlighted in \mred{red}), some of which are incorrect.}
    \label{fig:Ref}
\end{figure*}

\section{Experiments}
In this section, we conduct extensive experiments to validate the effectiveness of our proposed method. 

\subsection{Experimental Settings}
\paragraph{Benchmarks.} To evaluate visual reasoning capabilities of the model, we choose three high-resolution visual understanding datasets, including VisualProbe~\cite{arXiv:2025:Minio3}, V*~\cite{CVPR:2024:V*}, and HR-bench~\cite{AAAI:2025:HRBench}, all containing images with high resolutions ranging from 2K to 8K. In all datasets, the questions refer to small visual targets, making accurate region localization particularly challenging for models. Among them, VisualProbe features complex real-world scenes, especially in its hard category, where even human vision finds it difficult to locate the correct targets. These datasets require fine-grained visual understanding, thus  reflecting the advantages of the thinking with images paradigm. All results are reported using the pass@1 metric, which is a commonly adopted evaluation measure in reasoning tasks.

\paragraph{Training Details.} During the RL stage, we set the training batch size to 96, with a mini-batch size also of 96. For each prompt, 12 rollouts are generated, and the VLM agent is allowed a maximum of 5 interaction turns. Neither KL nor entropy regularization is applied. The threshold $\epsilon$ and hyper-parameter $\lambda$ are set to 0.5 and 0.2, respectively. 

\subsection{Main Results}

\paragraph{Comparison with Other Baselines.} We compare our method with three types of baselines, including \textit{(i)} frontier closed-source models: GPT-5~\cite{GPT5} and Gemini 2.5 Pro~\cite{gemini25}; \textit{(ii)} state-of-the-art open-source models: LLaVA-OneVision~\cite{Llava-onevision} and Ovis2.5~\cite{Ovis25}; and \textit{(iii)} visual reasoning models: SEAL~\cite{CVPR:2024:V*}, DyFo~\cite{CVPR:2025:Dyfo}, DeepEyes(v2)~\cite{arXiv:2025:DeepEyes,arXiv:2025:DeepEyesv2}, Thyme~\cite{arXiv:2025:Thyme} and Mini-o3~\cite{arXiv:2025:Minio3}. The comparison results between our method and the baselines on three high-resolution visual understanding benchmarks are presented in \Cref{tab:main}. It can be observed that our method, DRIM, achieves the best or comparable performance across all types of datasets. Compared with the second-best model, DRIM achieves significant improvements, achieving gains of 5.4\% on HR-bench 4K, 9.1\% on HR-bench 8K, and 2.5\% on VisualProbe overall. These results highlight the strong visual reasoning capability of our proposed method, and demonstrate the substantial improvements that the ``thinking with images'' paradigm brings to visual understanding.

\begin{table*}[t]
\renewcommand{\arraystretch}{1.1}
\setlength{\tabcolsep}{8pt}
\centering
\caption{\textbf{Ablation Results on visual understanding datasets}. Best and second best performance are highlighted in  \mred{red} and \mblue{blue}.}
\label{tab:ablation}
\begin{tabular}{c|ccc|ccc|c|cc}
\toprule
\multirow{2}{*}{\textbf{Models}} &
\multirow{2}{*}{\makecell{\textbf{Cold-start} \\ \textbf{SFT}}}  & \multicolumn{2}{c|}{\textbf{End-to-end RL}} &
\multicolumn{3}{c|}{\textbf{VisualProbe}}  &
\textbf{V*} & \multicolumn{2}{c}{\textbf{HR-Bench}} \\
&  & Acc Reward & Penalty & hard & medium & easy & overall & 4K & 8K \\
\midrule

\textsf{A} &  \XSolidBrush &  \CheckmarkBold & \CheckmarkBold & 20.8   & 23.1  & 55.3    & 68.6   & 74.0 & 69.5  \\

\textsf{B} & \CheckmarkBold & \XSolidBrush & \XSolidBrush & 30.2  & 34.3 & 60.3  &  78.9  &  75.3 & 72.1  \\
\textsf{C} & \CheckmarkBold & \CheckmarkBold & \XSolidBrush & \mblue{38.7}  & \mblue{47.4}  & \mblue{66.0}    & \mblue{90.1}    &  \mblue{81.8} & \mblue{80.8}  \\ \midrule

\textbf{DRIM} (Ours) & \CheckmarkBold & \CheckmarkBold & \CheckmarkBold & \mred{45.3} &	\mred{48.1} &	\mred{69.5}  &  \mred{92.2}  & \mred{83.3} & \mred{82.9} \\

\bottomrule
\end{tabular}
\end{table*}

\paragraph{Multi-scale Exploration.} To further understand why DRIM outperforms existing approaches, we qualitatively analyze the multi-turn reasoning trajectories produced by different models, as illustrated in \Cref{fig:Ref}. The trajectories reveal that DRIM is capable of performing multi-scale exploration, which is essential for solving fine-grained visual reasoning tasks. In particular, DRIM actively examines multiple spatial regions at different scales, gradually refining its visual focus. More importantly, the model exhibits a self-reflective behavior: when an intermediate observation is uninformative or misleading, DRIM corrects its reasoning direction by shifting the zoom-in region to a more plausible location (highlighted in \mblue{blue}). This reflective adjustment enables the model to converge to the correct answer even in complex scenes with numerous distractors. By contrast, RL training without our redundancy-penalty tends to generate oscillatory micro-adjustments around nearly identical regions, failing to expand the search area. This results in repeated zoom-ins on the same incorrect subregion, ultimately yielding a wrong answer despite multiple turns. Other baseline exhibits even shallower reasoning behaviors, frequently stopping after one or two turns and directly guessing an answer without thorough exploration. 

Overall, \Cref{fig:Ref} demonstrates that DRIM can perform multi-turn reliable reasoning in challenging environments. These qualitative findings corroborate our quantitative results, highlighting that encouraging multi-scale exploration and self-reflection during the reasoning process is crucial for ``thinking with images'' in visual reasoning tasks. 

\subsection{Ablation Study}
To further demonstrate the effectiveness of each component, we conduct the ablation study on three high-resolution visual understanding benchmarks, and the results are summarized in \Cref{tab:ablation}. All variants are trained using the same settings as the full model for fair comparisons. 

\paragraph{Effectiveness of Cold-start SFT.} To validate the effectiveness of cold-start SFT, we train Model \textsf{A}, which excludes the SFT stage while keeping all other components unchanged. From the results reported in \Cref{tab:ablation}, we can see that the performance of Model \textsf{A} drops substantially on both the VisualProbe and V* datasets. This degradation arises because the model struggles to acquire the ability to invoke tools and ``think with images'' without the SFT stage, making it difficult to perform fine-grained reasoning over complex visual scenes. In contrast, for datasets such as VisualProbe-Easy and HR-Bench 4K, many samples can be correctly answered without tool use, which explains why the Model \textsf{A} with only RL training still performs reasonably well on these benchmarks. Nevertheless, for challenging visual reasoning tasks, SFT plays an essential role by enabling the model to develop a multi-turn reasoning pattern that is crucial for multimodal reasoning. 

\paragraph{Effectiveness of RL.} The intention of introducing RL is to enhance the model’s ability for self-exploration and optimizing its policy during the reasoning process. To examine this, we train Model \textsf{B}, which performs SFT on tool call trajectories to teach the base model how to invoke tools, but does not apply RL to strengthen its reasoning capabilities. As shown in \Cref{tab:ablation}, Model \textsf{B} achieves notable improvements over Model \textsf{A} on most datasets, indicating that it can acquire basic tool-use abilities through SFT. However, its performance still lags far behind that of Models \textsf{C} and \textsf{D}, demonstrating that SFT by itself is insufficient and that RL is essential for achieving strong multi-turn reasoning. 

\paragraph{Effectiveness of Redundancy-Penalty.} We further train Model \textsf{C}, which incorporates both the SFT stage and an RL stage that uses only the accuracy reward. In \Cref{tab:ablation}, we observe that Model \textsf{C} achieves strong performance across multiple datasets and demonstrates robust reasoning ability even on challenging visual understanding tasks. Moreover, by introducing the redundancy-penalty term into the RL training, our DRIM attains additional performance gains on all of the high-resolution benchmarks, highlighting the effectiveness of encouraging multi-scale exploration beyond result-oriented optimization. 

\section{Conclusion}
In this paper, we presented \textbf{DRIM}, a model that enables \textbf{D}eep but \textbf{R}eliable multi-turn reasoning when thinking with \textbf{I}mages in its \textbf{M}ultimodal CoT. Motivated by the limitation that existing methods often struggle to reflect on and correct themselves when attempting incorrect reasoning trajectories, we introduce a training pipeline consisting of data construction, cold-start SFT and  RL. We construct a new multimodal reasoning dataset that satisfies \emph{high-difficulty} and \emph{verifiability}, which encourages models to invoke tools for visual reasoning. In the SFT stage, we synthesize tool call trajectories as cold-start data, guiding the model to multi-turn reasoning. In the RL stage, we introduce redundancy-penalized policy optimization, which incentivizes the model to develop a self-reflective reasoning pattern. Our DRIM achieves superior performance across multiple high-resolution visual understanding datasets, which is demonstrated by extensive experiments.

{
    \small
    \bibliographystyle{ieeenat_fullname}
    \bibliography{main}

@String(CVPR= {IEEE Conf. Comput. Vis. Pattern Recog.})

@String(ICCV= {Int. Conf. Comput. Vis.})

@String(AAAI = {AAAI})

@String(CVPR  = {CVPR})

@String(ICCV  = {ICCV})

@article{arXiv:2025:DeepEyes,
  title={DeepEyes: Incentivizing" Thinking with Images" via Reinforcement Learning},
  author={Zheng, Ziwei and Yang, Michael and Hong, Jack and Zhao, Chenxiao and Xu, Guohai and Yang, Le and Shen, Chao and Yu, Xing},
  journal={arXiv preprint arXiv:2505.14362},
  year={2025}
}

@article{arXiv:2025:DeepEyesv2,
  title={DeepEyesV2: Toward Agentic Multimodal Model},
  author={Hong, Jack and Zhao, Chenxiao and Zhu, ChengLin and Lu, Weiheng and Xu, Guohai and Yu, Xing},
  journal={arXiv preprint arXiv:2511.05271},
  year={2025}
}

@article{arXiv:2025:Minio3,
  title={Mini-o3: Scaling up reasoning patterns and interaction turns for visual search},
  author={Lai, Xin and Li, Junyi and Li, Wei and Liu, Tao and Li, Tianjian and Zhao, Hengshuang},
  journal={arXiv preprint arXiv:2509.07969},
  year={2025}
}

@article{Llava-onevision,
  title={Llava-onevision: Easy visual task transfer},
  author={Li, Bo and Zhang, Yuanhan and Guo, Dong and Zhang, Renrui and Li, Feng and Zhang, Hao and Zhang, Kaichen and Zhang, Peiyuan and Li, Yanwei and Liu, Ziwei and others},
  journal={arXiv preprint arXiv:2408.03326},
  year={2024}
}

@misc{llava-next,
  title={Llavanext: Improved reasoning, ocr, and world knowledge},
  author={Liu, Haotian and Li, Chunyuan and Li, Yuheng and Li, Bo and Zhang, Yuanhan and Shen, Sheng and Lee, Yong Jae},
  year={2024},
  howpublished = {\url{https://llava-vl.github.io/
blog/2024-01-30-llava-next/}}
}

@article{Qwen25-VL,
  title={Qwen2. 5-vl technical report},
  author={Bai, Shuai and Chen, Keqin and Liu, Xuejing and Wang, Jialin and Ge, Wenbin and Song, Sibo and Dang, Kai and Wang, Peng and Wang, Shijie and Tang, Jun and others},
  journal={arXiv preprint arXiv:2502.13923},
  year={2025}
}

@article{Internvl3,
  title={Internvl3: Exploring advanced training and test-time recipes for open-source multimodal models},
  author={Zhu, Jinguo and Wang, Weiyun and Chen, Zhe and Liu, Zhaoyang and Ye, Shenglong and Gu, Lixin and Tian, Hao and Duan, Yuchen and Su, Weijie and Shao, Jie and others},
  journal={arXiv preprint arXiv:2504.10479},
  year={2025}
}

@article{GLM-41V,
  title={GLM-4.1 V-Thinking: Towards Versatile Multimodal Reasoning with Scalable Reinforcement Learning},
  author={Team, V and Hong, W and Yu, W and Gu, X and Wang, G and Gan, G and Tang, H and Cheng, J and Qi, J and Ji, J and others},
  journal={arXiv preprint arXiv:2507.01006},
  year={2025}
}

@article{Keye-VL,
  title={Kwai Keye-VL Technical Report},
  author={Team, Kwai Keye and Yang, Biao and Wen, Bin and Liu, Changyi and Chu, Chenglong and Song, Chengru and Rao, Chongling and Yi, Chuan and Li, Da and Zang, Dunju and others},
  journal={arXiv preprint arXiv:2507.01949},
  year={2025}
}

@article{Ovis25,
  title={Ovis2. 5 technical report},
  author={Lu, Shiyin and Li, Yang and Xia, Yu and Hu, Yuwei and Zhao, Shanshan and Ma, Yanqing and Wei, Zhichao and Li, Yinglun and Duan, Lunhao and Zhao, Jianshan and others},
  journal={arXiv preprint arXiv:2508.11737},
  year={2025}
}

@article{arXiv:2023:Zhang,
  title={Multimodal chain-of-thought reasoning in language models},
  author={Zhang, Zhuosheng and Zhang, Aston and Li, Mu and Zhao, Hai and Karypis, George and Smola, Alex},
  journal={arXiv preprint arXiv:2302.00923},
  year={2023}
}

@inproceedings{ACL:2025:He,
    title = "{MMB}oundary: Advancing {MLLM} Knowledge Boundary Awareness through Reasoning Step Confidence Calibration",
    author = "He, Zhitao  and
      Polisetty, Sandeep  and
      Fan, Zhiyuan  and
      Huang, Yuchen  and
      Wu, Shujin  and
      Fung, Yi R.",
    booktitle = "Proceedings of the 63rd Annual Meeting of the Association for Computational Linguistics (Volume 1: Long Papers)",
    year = "2025",
    pages = "16427--16444",
}

@article{arXiv:2025:Shen,
  title={Satori-r1: Incentivizing multimodal reasoning with spatial grounding and verifiable rewards},
  author={Shen, Chuming and Wei, Wei and Qu, Xiaoye and Cheng, Yu},
  journal={arXiv preprint arXiv:2505.19094},
  year={2025}
}

@article{Mathvista,
  title={Mathvista: Evaluating mathematical reasoning of foundation models in visual contexts},
  author={Lu, Pan and Bansal, Hritik and Xia, Tony and Liu, Jiacheng and Li, Chunyuan and Hajishirzi, Hannaneh and Cheng, Hao and Chang, Kai-Wei and Galley, Michel and Gao, Jianfeng},
  journal={arXiv preprint arXiv:2310.02255},
  year={2023}
}

@inproceedings{NeurIPS:2022:Wei,
 author = {Wei, Jason and Wang, Xuezhi and Schuurmans, Dale and Bosma, Maarten and ichter, brian and Xia, Fei and Chi, Ed and Le, Quoc V and Zhou, Denny},
 booktitle = {Advances in Neural Information Processing Systems},
 pages = {24824--24837},
 title = {Chain-of-Thought Prompting Elicits Reasoning in Large Language Models},
 year = {2022}
}

@inproceedings{NeurIPS:2022:Kojima,
 author = {Kojima, Takeshi and Gu, Shixiang (Shane) and Reid, Machel and Matsuo, Yutaka and Iwasawa, Yusuke},
 booktitle = {Advances in Neural Information Processing Systems},
 pages = {22199--22213},
 title = {Large Language Models are Zero-Shot Reasoners},
 year = {2022}
}

@inproceedings{NeurIPS:2024:Hu,
 author = {Hu, Yushi and Shi, Weijia and Fu, Xingyu and Roth, Dan and Ostendorf, Mari and Zettlemoyer, Luke and Smith, Noah A and Krishna, Ranjay},
 booktitle = {Advances in Neural Information Processing Systems},
 pages = {139348--139379},
 title = {Visual Sketchpad: Sketching as a Visual Chain of Thought for Multimodal Language Models},
 year = {2024}
}

@article{arXiv:2025:Thyme,
  title={Thyme: Think beyond images},
  author={Zhang, Yi-Fan and Lu, Xingyu and Yin, Shukang and Fu, Chaoyou and Chen, Wei and Hu, Xiao and Wen, Bin and Jiang, Kaiyu and Liu, Changyi and Zhang, Tianke and others},
  journal={arXiv preprint arXiv:2508.11630},
  year={2025}
}

@InProceedings{CVPR:2024:V*,
    author    = {Wu, Penghao and Xie, Saining},
    title     = {V?: Guided Visual Search as a Core Mechanism in Multimodal LLMs},
    booktitle = {Proceedings of the IEEE/CVF Conference on Computer Vision and Pattern Recognition},
    year      = {2024},
    pages     = {13084-13094}
}

@InProceedings{CVPR:2025:Dyfo,
    author    = {Li, Geng and Xu, Jinglin and Zhao, Yunzhen and Peng, Yuxin},
    title     = {DyFo: A Training-Free Dynamic Focus Visual Search for Enhancing LMMs in Fine-Grained Visual Understanding},
    booktitle = {Proceedings of the IEEE/CVF Conference on Computer Vision and Pattern Recognition},
    year      = {2025},
    pages     = {9098-9108}
}

@misc{openaio3,
  title={Thinking with images},
  author={OpenAI},
  year={2025},
  howpublished = {\url{https://openai.com/index/thinking-with-images/}}
}

@article{LARKIN198765,
title = {Why a Diagram is (Sometimes) Worth Ten Thousand Words},
journal = {Cognitive Science},
pages = {65-100},
year = {1987},
author = {Jill H. Larkin and Herbert A. Simon}
}

@misc{openaio1,
  title={Introducing openai o1},
  author={OpenAI},
  year={2024},
  howpublished = {\url{https://openai.com/o1/}}
}

@misc{GPT5,
  title={GPT-5 System Card},
  author={OpenAI},
  year={2025},
  howpublished = {\url{https://cdn.openai.com/gpt-5-system-card.pdf}}
}

@article{gemini25,
  title={Gemini 2.5: Pushing the frontier with advanced reasoning, multimodality, long context, and next generation agentic capabilities},
  author={Comanici, Gheorghe and Bieber, Eric and Schaekermann, Mike and Pasupat, Ice and Sachdeva, Noveen and Dhillon, Inderjit and Blistein, Marcel and Ram, Ori and Zhang, Dan and Rosen, Evan and others},
  journal={arXiv preprint arXiv:2507.06261},
  year={2025}
}

@article{DeepSeek-r1,
  title={Deepseek-r1: Incentivizing reasoning capability in llms via reinforcement learning},
  author={Guo, Daya and Yang, Dejian and Zhang, Haowei and Song, Junxiao and Zhang, Ruoyu and Xu, Runxin and Zhu, Qihao and Ma, Shirong and Wang, Peiyi and Bi, Xiao and others},
  journal={arXiv preprint arXiv:2501.12948},
  year={2025}
}

@inproceedings{verl,
author = {Sheng, Guangming and Zhang, Chi and Ye, Zilingfeng and Wu, Xibin and Zhang, Wang and Zhang, Ru and Peng, Yanghua and Lin, Haibin and Wu, Chuan},
title = {HybridFlow: A Flexible and Efficient RLHF Framework},
year = {2025},
booktitle = {Proceedings of the Twentieth European Conference on Computer Systems},
pages = {1279–1297}
}

@article{GRPO,
  title={Deepseekmath: Pushing the limits of mathematical reasoning in open language models},
  author={Shao, Zhihong and Wang, Peiyi and Zhu, Qihao and Xu, Runxin and Song, Junxiao and Bi, Xiao and Zhang, Haowei and Zhang, Mingchuan and Li, YK and Wu, Yang and others},
  journal={arXiv preprint arXiv:2402.03300},
  year={2024}
}

@article{reinforce++,
  title={Reinforce++: A simple and efficient approach for aligning large language models},
  author={Hu, Jian},
  journal={arXiv preprint arXiv:2501.03262},
  year={2025}
}

@article{RLOO,
  title={Back to basics: Revisiting reinforce style optimization for learning from human feedback in llms},
  author={Ahmadian, Arash and Cremer, Chris and Gall{\'e}, Matthias and Fadaee, Marzieh and Kreutzer, Julia and Pietquin, Olivier and {\"U}st{\"u}n, Ahmet and Hooker, Sara},
  journal={arXiv preprint arXiv:2402.14740},
  year={2024}
}

@article{dapo,
  title={Dapo: An open-source llm reinforcement learning system at scale},
  author={Yu, Qiying and Zhang, Zheng and Zhu, Ruofei and Yuan, Yufeng and Zuo, Xiaochen and Yue, Yu and Dai, Weinan and Fan, Tiantian and Liu, Gaohong and Liu, Lingjun and others},
  journal={arXiv preprint arXiv:2503.14476},
  year={2025}
}

@inproceedings{AAAI:2025:HRBench,
author = {Wang, Wenbin and Ding, Liang and Zeng, Minyan and Zhou, Xiabin and Shen, Li and Luo, Yong and Yu, Wei and Tao, Dacheng},
title = {Divide, conquer and combine: a training-free framework for high-resolution image perception in multimodal large language models},
year = {2025},
booktitle = {Proceedings of the Thirty-Ninth AAAI Conference on Artificial Intelligence and Thirty-Seventh Conference on Innovative Applications of Artificial Intelligence and Fifteenth Symposium on Educational Advances in Artificial Intelligence}
}

@InProceedings{NeurIPS:2021:CLIP,
  title = 	 {Learning Transferable Visual Models From Natural Language Supervision},
  author =       {Radford, Alec and Kim, Jong Wook and Hallacy, Chris and Ramesh, Aditya and Goh, Gabriel and Agarwal, Sandhini and Sastry, Girish and Askell, Amanda and Mishkin, Pamela and Clark, Jack and Krueger, Gretchen and Sutskever, Ilya},
  booktitle = 	 {Proceedings of the 38th International Conference on Machine Learning},
  pages = 	 {8748--8763},
  year = 	 {2021}
}

@InProceedings{ICCV:2023:Siglip,
    author    = {Zhai, Xiaohua and Mustafa, Basil and Kolesnikov, Alexander and Beyer, Lucas},
    title     = {Sigmoid Loss for Language Image Pre-Training},
    booktitle = {Proceedings of the IEEE/CVF International Conference on Computer Vision},
    year      = {2023},
    pages     = {11975-11986}
}

@InProceedings{ICML:2022:BLIP,
  title = 	 {{BLIP}: Bootstrapping Language-Image Pre-training for Unified Vision-Language Understanding and Generation},
  author =       {Li, Junnan and Li, Dongxu and Xiong, Caiming and Hoi, Steven},
  booktitle = 	 {Proceedings of the 39th International Conference on Machine Learning},
  pages = 	 {12888--12900},
  year = 	 {2022}
}

@InProceedings{ICML:2023:BLIP2,
  title = 	 {{BLIP}-2: Bootstrapping Language-Image Pre-training with Frozen Image Encoders and Large Language Models},
  author =       {Li, Junnan and Li, Dongxu and Savarese, Silvio and Hoi, Steven},
  booktitle = 	 {Proceedings of the 40th International Conference on Machine Learning},
  pages = 	 {19730--19742},
  year = 	 {2023}
}

@inproceedings{NeurIPS:2022:Flamingo,
 author = {Alayrac, Jean-Baptiste and Donahue, Jeff and Luc, Pauline and Miech, Antoine and Barr, Iain and Hasson, Yana and Lenc, Karel and Mensch, Arthur and Millican, Katherine and Reynolds, Malcolm and Ring, Roman and Rutherford, Eliza and Cabi, Serkan and Han, Tengda and Gong, Zhitao and Samangooei, Sina and Monteiro, Marianne and Menick, Jacob L and Borgeaud, Sebastian and Brock, Andy and Nematzadeh, Aida and Sharifzadeh, Sahand and Bi\'{n}kowski, Miko\l aj and Barreira, Ricardo and Vinyals, Oriol and Zisserman, Andrew and Simonyan, Kar\'{e}n},
 booktitle = {Advances in Neural Information Processing Systems},
 pages = {23716--23736},
 title = {Flamingo: a Visual Language Model for Few-Shot Learning},
 year = {2022}
}

@inproceedings{NeurIPS:2023:LLaVA,
 author = {Liu, Haotian and Li, Chunyuan and Wu, Qingyang and Lee, Yong Jae},
 booktitle = {Advances in Neural Information Processing Systems},
 pages = {34892--34916},
 title = {Visual Instruction Tuning},
 year = {2023}
}

@inproceedings{ViT,
title={An Image is Worth 16x16 Words: Transformers for Image Recognition at Scale},
author={Alexey Dosovitskiy and Lucas Beyer and Alexander Kolesnikov and Dirk Weissenborn and Xiaohua Zhai and Thomas Unterthiner and Mostafa Dehghani and Matthias Minderer and Georg Heigold and Sylvain Gelly and Jakob Uszkoreit and Neil Houlsby},
booktitle={International Conference on Learning Representations},
year={2021}
}

@InProceedings{LLaVA15,
    author    = {Liu, Haotian and Li, Chunyuan and Li, Yuheng and Lee, Yong Jae},
    title     = {Improved Baselines with Visual Instruction Tuning},
    booktitle = {Proceedings of the IEEE/CVF Conference on Computer Vision and Pattern Recognition},
    year      = {2024},
    pages     = {26296-26306}
}

@InProceedings{internvl,
    author    = {Chen, Zhe and Wu, Jiannan and Wang, Wenhai and Su, Weijie and Chen, Guo and Xing, Sen and Zhong, Muyan and Zhang, Qinglong and Zhu, Xizhou and Lu, Lewei and Li, Bin and Luo, Ping and Lu, Tong and Qiao, Yu and Dai, Jifeng},
    title     = {InternVL: Scaling up Vision Foundation Models and Aligning for Generic Visual-Linguistic Tasks},
    booktitle = {Proceedings of the IEEE/CVF Conference on Computer Vision and Pattern Recognition},
    year      = {2024},
    pages     = {24185-24198}
}

@article{internvl15,
  title={How far are we to gpt-4v? closing the gap to commercial multimodal models with open-source suites},
  author={Chen, Zhe and Wang, Weiyun and Tian, Hao and Ye, Shenglong and Gao, Zhangwei and Cui, Erfei and Tong, Wenwen and Hu, Kongzhi and Luo, Jiapeng and Ma, Zheng and others},
  journal={Science China Information Sciences},
  pages={220101},
  year={2024}
}

@article{internvl25,
  title={Expanding performance boundaries of open-source multimodal models with model, data, and test-time scaling},
  author={Chen, Zhe and Wang, Weiyun and Cao, Yue and Liu, Yangzhou and Gao, Zhangwei and Cui, Erfei and Zhu, Jinguo and Ye, Shenglong and Tian, Hao and Liu, Zhaoyang and others},
  journal={arXiv preprint arXiv:2412.05271},
  year={2024}
}

@article{internvl35,
  title={Internvl3. 5: Advancing open-source multimodal models in versatility, reasoning, and efficiency},
  author={Wang, Weiyun and Gao, Zhangwei and Gu, Lixin and Pu, Hengjun and Cui, Long and Wei, Xingguang and Liu, Zhaoyang and Jing, Linglin and Ye, Shenglong and Shao, Jie and others},
  journal={arXiv preprint arXiv:2508.18265},
  year={2025}
}

@misc{qwenvl,
      title={Qwen-VL: A Versatile Vision-Language Model for Understanding, Localization, Text Reading, and Beyond}, 
      author={Jinze Bai and Shuai Bai and Shusheng Yang and Shijie Wang and Sinan Tan and Peng Wang and Junyang Lin and Chang Zhou and Jingren Zhou},
      year={2023},
      eprint={2308.12966},
      archivePrefix={arXiv},
      primaryClass={cs.CV},
      url={https://arxiv.org/abs/2308.12966}, 
}

@article{qwen2vl,
  title={Qwen2-vl: Enhancing vision-language model's perception of the world at any resolution},
  author={Wang, Peng and Bai, Shuai and Tan, Sinan and Wang, Shijie and Fan, Zhihao and Bai, Jinze and Chen, Keqin and Liu, Xuejing and Wang, Jialin and Ge, Wenbin and others},
  journal={arXiv preprint arXiv:2409.12191},
  year={2024}
}

@article{ovis,
  title={Ovis: Structural embedding alignment for multimodal large language model},
  author={Lu, Shiyin and Li, Yang and Chen, Qing-Guo and Xu, Zhao and Luo, Weihua and Zhang, Kaifu and Ye, Han-Jia},
  journal={arXiv preprint arXiv:2405.20797},
  year={2024}
}

@article{seed15vl,
  title={Seed1. 5-vl technical report},
  author={Guo, Dong and Wu, Faming and Zhu, Feida and Leng, Fuxing and Shi, Guang and Chen, Haobin and Fan, Haoqi and Wang, Jian and Jiang, Jianyu and Wang, Jiawei and others},
  journal={arXiv preprint arXiv:2505.07062},
  year={2025}
}

@article{arXiv:2025:Huang,
  title={Visualtoolagent (vista): A reinforcement learning framework for visual tool selection},
  author={Huang, Zeyi and Ji, Yuyang and Rajan, Anirudh Sundara and Cai, Zefan and Xiao, Wen and Wang, Haohan and Hu, Junjie and Lee, Yong Jae},
  journal={arXiv preprint arXiv:2505.20289},
  year={2025}
}

@inproceedings{EMNLP:2025:zoomeye,
    title = "{Z}oom{E}ye: Enhancing Multimodal {LLM}s with Human-Like Zooming Capabilities through Tree-Based Image Exploration",
    author = "Shen, Haozhan  and
      Zhao, Kangjia  and
      Zhao, Tiancheng  and
      Xu, Ruochen  and
      Zhang, Zilun  and
      Zhu, Mingwei  and
      Yin, Jianwei",
    booktitle = "Proceedings of the 2025 Conference on Empirical Methods in Natural Language Processing",
    year = "2025",
    pages = "6613--6629"
}

@article{arXiv:2023:Yang,
  title={Mm-react: Prompting chatgpt for multimodal reasoning and action},
  author={Yang, Zhengyuan and Li, Linjie and Wang, Jianfeng and Lin, Kevin and Azarnasab, Ehsan and Ahmed, Faisal and Liu, Zicheng and Liu, Ce and Zeng, Michael and Wang, Lijuan},
  journal={arXiv preprint arXiv:2303.11381},
  year={2023}
}

@article{arXiv:2025:Cheng,
  title={Visual thoughts: A unified perspective of understanding multimodal chain-of-thought},
  author={Cheng, Zihui and Chen, Qiguang and Xu, Xiao and Wang, Jiaqi and Wang, Weiyun and Fei, Hao and Wang, Yidong and Wang, Alex Jinpeng and Chen, Zhi and Che, Wanxiang and others},
  journal={arXiv preprint arXiv:2505.15510},
  year={2025}
}

@inproceedings{ECCV:2024:Liu,
  title={Llava-plus: Learning to use tools for creating multimodal agents},
  author={Liu, Shilong and Cheng, Hao and Liu, Haotian and Zhang, Hao and Li, Feng and Ren, Tianhe and Zou, Xueyan and Yang, Jianwei and Su, Hang and Zhu, Jun and others},
  booktitle={European conference on computer vision},
  pages={126--142},
  year={2024},
  organization={Springer}
}

@article{arXiv:2025:Bai,
  title={Multi-Step Visual Reasoning with Visual Tokens Scaling and Verification},
  author={Bai, Tianyi and Hu, Zengjie and Sun, Fupeng and Qiu, Jiantao and Jiang, Yizhen and He, Guangxin and Zeng, Bohan and He, Conghui and Yuan, Binhang and Zhang, Wentao},
  journal={arXiv preprint arXiv:2506.07235},
  year={2025}
}

@article{corr:2024:Shao,
  title={Visual cot: Unleashing chain-of-thought reasoning in multi-modal language models},
  author={Shao, Hao and Qian, Shengju and Xiao, Han and Song, Guanglu and Zong, Zhuofan and Wang, Letian and Liu, Yu and Li, Hongsheng},
  journal={CoRR},
  year={2024}
}

@article{arXiv:2025:Zhang,
  title={Cmmcot: Enhancing complex multi-image comprehension via multi-modal chain-of-thought and memory augmentation},
  author={Zhang, Guanghao and Zhong, Tao and Xia, Yan and Yu, Zhelun and Li, Haoyuan and He, Wanggui and Shu, Fangxun and Liu, Mushui and She, Dong and Wang, Yi and others},
  journal={arXiv preprint arXiv:2503.05255},
  year={2025}
}

@article{arXiv:2025:Liu,
  title={VisionReasoner: Unified Visual Perception and Reasoning via Reinforcement Learning},
  author={Liu, Yuqi and Qu, Tianyuan and Zhong, Zhisheng and Peng, Bohao and Liu, Shu and Yu, Bei and Jia, Jiaya},
  journal={arXiv preprint arXiv:2505.12081},
  year={2025}
}

@article{arXiv:2025:Zhu,
  title={Active-O3: Empowering Multimodal Large Language Models with Active Perception via GRPO},
  author={Zhu, Muzhi and Zhong, Hao and Zhao, Canyu and Du, Zongze and Huang, Zheng and Liu, Mingyu and Chen, Hao and Zou, Cheng and Chen, Jingdong and Yang, Ming and others},
  journal={arXiv preprint arXiv:2505.21457},
  year={2025}
}

@article{arXiv:2025:Zhang:B,
  title={Chain-of-Focus: Adaptive Visual Search and Zooming for Multimodal Reasoning via RL},
  author={Zhang, Xintong and Gao, Zhi and Zhang, Bofei and Li, Pengxiang and Zhang, Xiaowen and Liu, Yang and Yuan, Tao and Wu, Yuwei and Jia, Yunde and Zhu, Song-Chun and others},
  journal={arXiv preprint arXiv:2505.15436},
  year={2025}
}

@article{arXiv:2025:Su,
  title={Pixel reasoner: Incentivizing pixel-space reasoning with curiosity-driven reinforcement learning},
  author={Su, Alex and Wang, Haozhe and Ren, Weiming and Lin, Fangzhen and Chen, Wenhu},
  journal={arXiv preprint arXiv:2505.15966},
  year={2025}
}

@inproceedings{NeurIPS:2023:Zhu,
 author = {Zhu, Wanrong and Hessel, Jack and Awadalla, Anas and Gadre, Samir Yitzhak and Dodge, Jesse and Fang, Alex and Yu, Youngjae and Schmidt, Ludwig and Wang, William Yang and Choi, Yejin},
 booktitle = {Advances in Neural Information Processing Systems},
 pages = {8958--8974},
 title = {Multimodal C4: An Open, Billion-scale Corpus of Images Interleaved with Text},
 year = {2023}
}

@InProceedings{CVPR:2016:Zhu,
author = {Zhu, Yuke and Groth, Oliver and Bernstein, Michael and Fei-Fei, Li},
title = {Visual7W: Grounded Question Answering in Images},
booktitle = {Proceedings of the IEEE Conference on Computer Vision and Pattern Recognition (CVPR)},
year = {2016}
}

@InProceedings{SA1B,
    author    = {Kirillov, Alexander and Mintun, Eric and Ravi, Nikhila and Mao, Hanzi and Rolland, Chloe and Gustafson, Laura and Xiao, Tete and Whitehead, Spencer and Berg, Alexander C. and Lo, Wan-Yen and Dollar, Piotr and Girshick, Ross},
    title     = {Segment Anything},
    booktitle = {Proceedings of the IEEE/CVF International Conference on Computer Vision (ICCV)},
    year      = {2023},
    pages     = {4015-4026}
}

@inproceedings{FineWeb,
    title = "{VISA}: Retrieval Augmented Generation with Visual Source Attribution",
    author = "Ma, Xueguang  and
      Zhuang, Shengyao  and
      Koopman, Bevan  and
      Zuccon, Guido  and
      Chen, Wenhu  and
      Lin, Jimmy",
    editor = "Che, Wanxiang  and
      Nabende, Joyce  and
      Shutova, Ekaterina  and
      Pilehvar, Mohammad Taher",
    booktitle = "Proceedings of the 63rd Annual Meeting of the Association for Computational Linguistics (Volume 1: Long Papers)",
    year = "2025",
    pages = "30154--30169"
}

@article{starqa,
  title={Star: A benchmark for situated reasoning in real-world videos},
  author={Wu, Bo and Yu, Shoubin and Chen, Zhenfang and Tenenbaum, Joshua B and Gan, Chuang},
  journal={arXiv preprint arXiv:2405.09711},
  year={2024}
}
}

\maketitlesupplementaryoc

Our supplementary material provides more details about our method, which can be summarized as follows: 
\begin{itemize}
    \item We provide the prompts used in our training scheme in Section~\ref{supp:prompt}. 
    \item We provide the data composition in the data construction stage in Section~\ref{supp:data}. 
    \item We provide more illustrations of multi-turn reasoning in Section~\ref{supp:case}
    \item We provide the future directions for ``thinking with images'' in Section~\ref{supp:fut}. 
\end{itemize}

\section{Prompts}
\label{supp:prompt}
In this section, we provide the prompt templates used during training and evaluation, including a system prompt, a user prompt, and a assistant prompt. 

\begin{promptblock}{SYSTEM PROMPT}
\begin{PromptVerb}
You are a helpful assistant.

# Context
In each turn, new images might be provided as a result of your tool calls. The images are numbered sequentially starting from 1. You can refer to any image that has appeared so far in the conversation using its `image_idx`.

# Tools
You may call one or more functions to assist with the user query.
You are provided with function signatures within <tools></tools> XML tags:
<tools>
{
  "type":"function",
  "function":{
    "name":"image_zoom_in_tool",
    "description":"Zoom in on a specific region of an image by cropping it. The new cropped image will be available in the next turn.",
    "parameters":{
      "type":"object",
      "properties":{
        "image_idx":{
          "type":"integer",
          "description":"The 1-based index of the image to perform the zoom-in operation on. The available images are provided and numbered in the user's prompt."
        },
        "bbox_2d":{
          "type":"string",
          "description":"The bounding box of the region to zoom in, as a string '<box>(x1,y1),(x2,y2)</box>' in relative coordinates (0.0 to 1.0) for the selected image, where (x1, y1) is the top-left corner and (x2, y2) is the bottom-right corner."
        },
        "label":{
          "type":"string",
          "description":"The name or label of the object in the specified bounding box (optional)."
        }
      },
      "required":["image_idx","bbox_2d"]
    }
  }
}
</tools>

# How to call a tool
Return a json object with function name and arguments within <tool_call></tool_call> XML tags:
<tool_call>
{"name": <function-name>, "arguments": <args-json-object>}
</tool_call>
\end{PromptVerb}
\end{promptblock}

\begin{promptblock}{USER PROMPT}
\begin{PromptVerb}
Image 1:
Question: {question}
\end{PromptVerb}
\end{promptblock}

\begin{promptblock}{ASSISTANT PROMPT}
\begin{PromptVerb}
<image> {image_zoom_in} </image>
<tool_response>
Image {new_idx} (cropped from Image {image_idx}) is provided.
</tool_response>
\end{PromptVerb}
\end{promptblock}

\section{Data Composition}
\label{supp:data}
In this section, we detail the data collection strategy adopted during the data construction stage. 

\subsection{Cold-start SFT Data}
During the SFT stage, our cold-start data are composed of two parts: a \textbf{H}igh-\textbf{R}esolution \textbf{V}isual Hard (\textbf{HRV}-Hard) dataset we constructed, and publicly released visual datasets. The data distribution is summarized as follows: 
\begin{itemize}
    \item \textbf{High-Resolution Visual Hard Dataset (47k samples):} All images are selected at a resolution of $1600\times 1600$ pixels, which ensures sufficiently high visual complexity for fine-grained analysis. The dataset includes the high-resolution images we curated, covering natural scenes, stylized images, and several other categories, primarily collected from publicly available websites. In addition, we incorporate a subset of visual data sampled from the open-source \texttt{mmc4} dataset~\cite{NeurIPS:2023:Zhu}. 
    \item \textbf{Open-Source Visual Dataset (22k samples):} We include datasets constructed by two publicly available works: PixelReasoner~\cite{arXiv:2025:Su} and Visual7W~\cite{CVPR:2016:Zhu}. For PixelReasoner, its dataset is built upon three publicly available sources, including SA1B~\cite{SA1B}, FineWeb~\cite{FineWeb} and STARQA~\cite{starqa}. For Visual7W, its dataset is constructed based on the Microsoft COCO dataset. 
\end{itemize}

\paragraph{Trajectory Synthesis} Based on the collected high-resolution visual dataset, we synthesize multimodal reasoning trajectories as cold-start data. In this work, we do not adopt trajectories provided by existing open-source efforts. Instead, we construct an automated pipeline that leverages frontier visual reasoning models to synthesize these trajectories. Specifically, for our HRV-Hard dataset, we employ Gemini and o4-mini to autonomously zoom in on regions of interest and generate QA pairs. For the open-source visual datasets, we directly use their original QA pairs. Finally, we utilize o4-mini to synthesize multimodal reasoning trajectories, which serve as our cold-start data. 

\subsection{RL Data}
During the RL training stage, our dataset consists of two main components, as outlined below:
\begin{itemize}
    \item \textbf{HRV-Hard (30k samples):} The main portion of the RL training data is sampled from our HRV-Hard dataset, from which we select 30k samples. The data cover a wide range of sources, including general natural images, stylized images, table images, and visual evidence documents. Nearly all training samples require the model to perform complex visual reasoning, thereby encouraging the model to think with images within its MCoT. 
    \item \textbf{VisualProbe (5k samples):} We also incorporate recently released open-source datasets as part of the RL training data. VisualProbe~\cite{arXiv:2025:Minio3} is a high-quality visual reasoning dataset primarily composed of natural images. The regions associated with the correct answers are often extremely difficult to localize, which effectively incentivizes the model’s multi-turn visual grounding and reasoning capabilities during training.
\end{itemize}

\section{More Illustrations of Multi-turn Reasoning}
\label{supp:case}
In this section, we provide more cases to demonstrate that DRIM can reflect and self-correct during the reasoning process (highlighted in \mblue{blue}), thereby localizing the correct region and producing the final answer (highlighted in \mred{red}).

\begin{figure}[h]
\begin{center}
\centerline{\includegraphics[width=.64\linewidth]{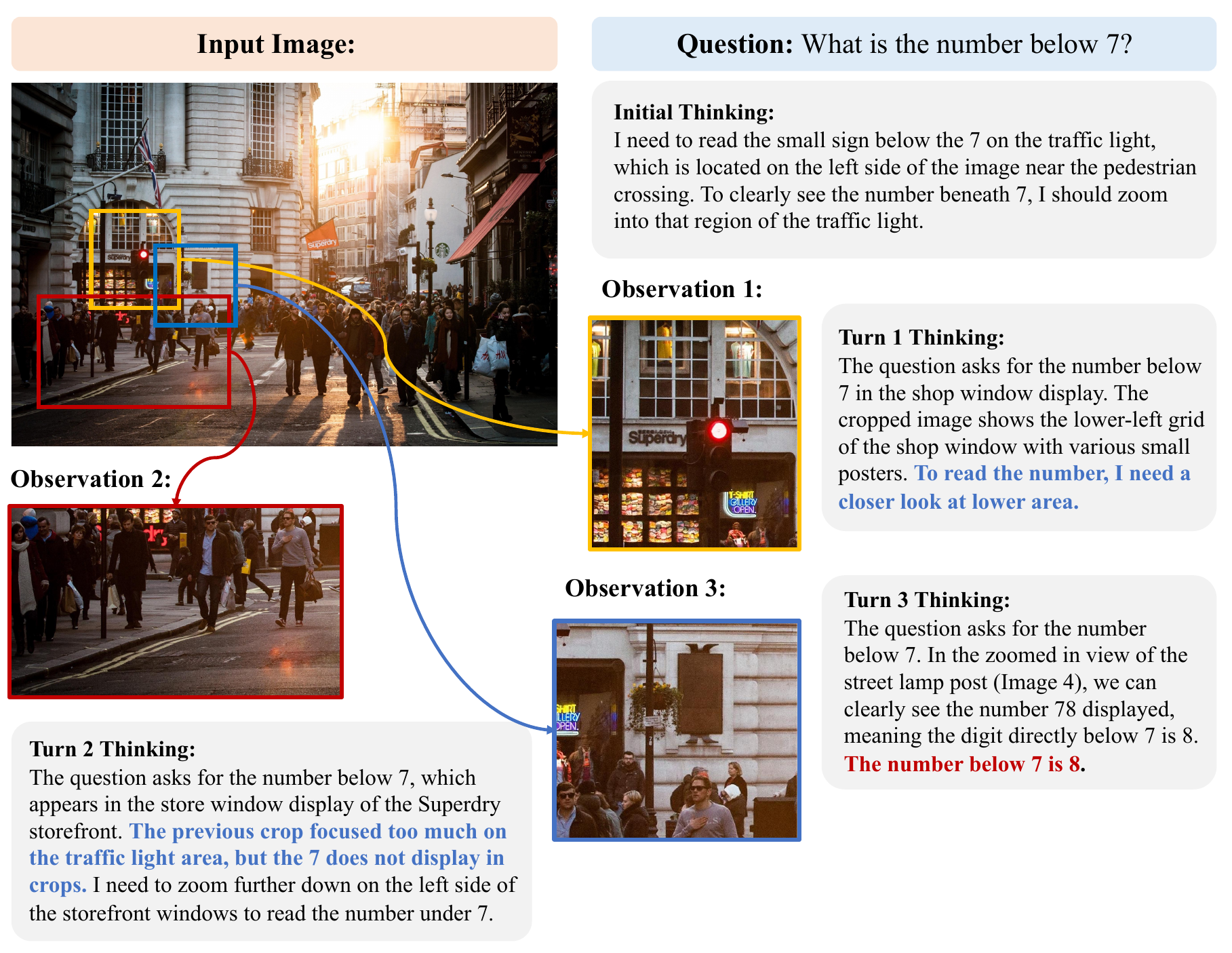}}
\caption{Case I: illustration of multi-turn reasoning}
\label{fig:case11}
\end{center}
\end{figure}

\begin{figure}[h]
\begin{center}
\centerline{\includegraphics[width=.64\linewidth]{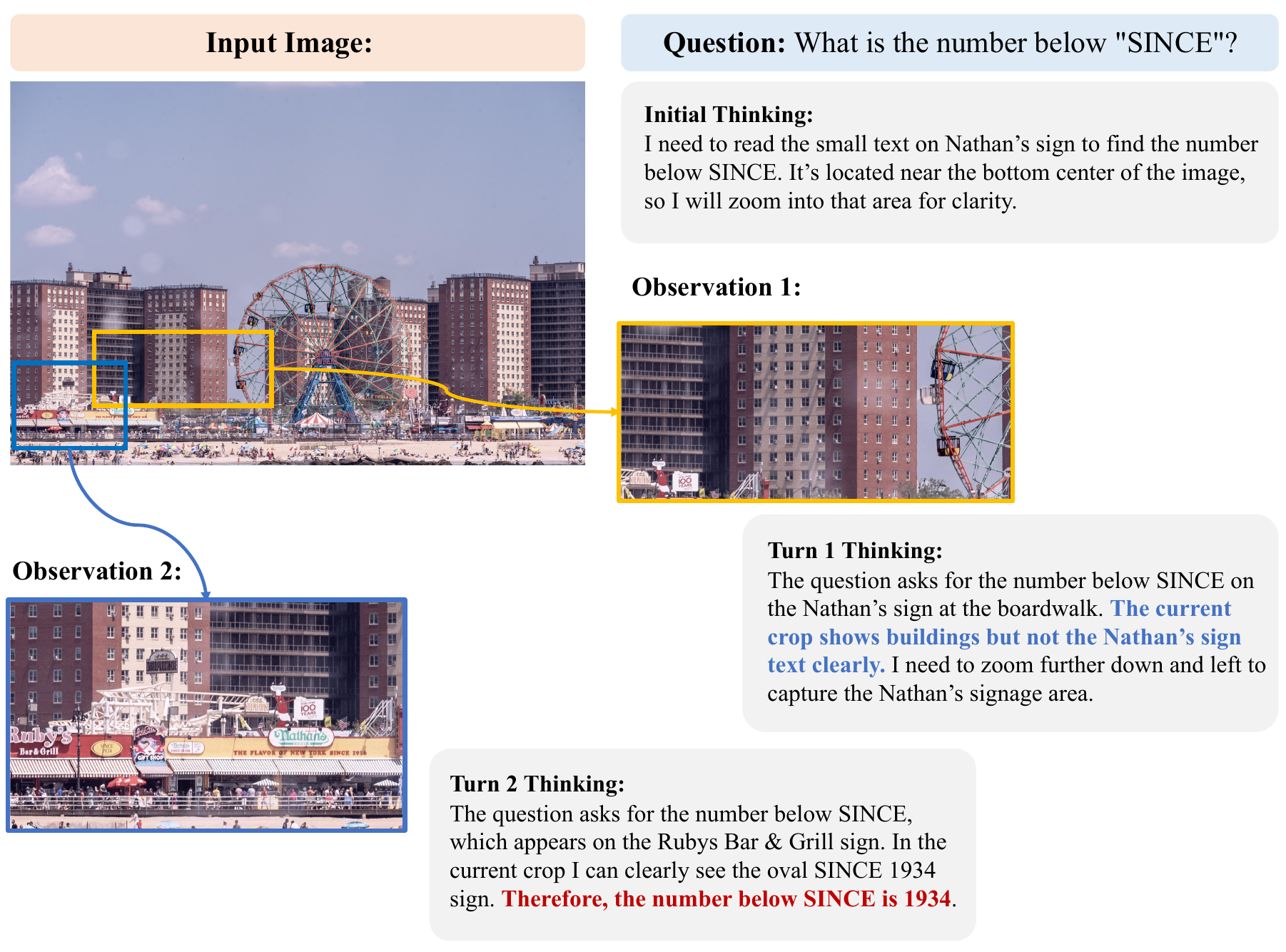}}
\caption{Case II: illustration of multi-turn reasoning}
\label{fig:case12}
\end{center}
\end{figure}

\begin{figure}[t]
\begin{center}
\centerline{\includegraphics[width=.96\linewidth]{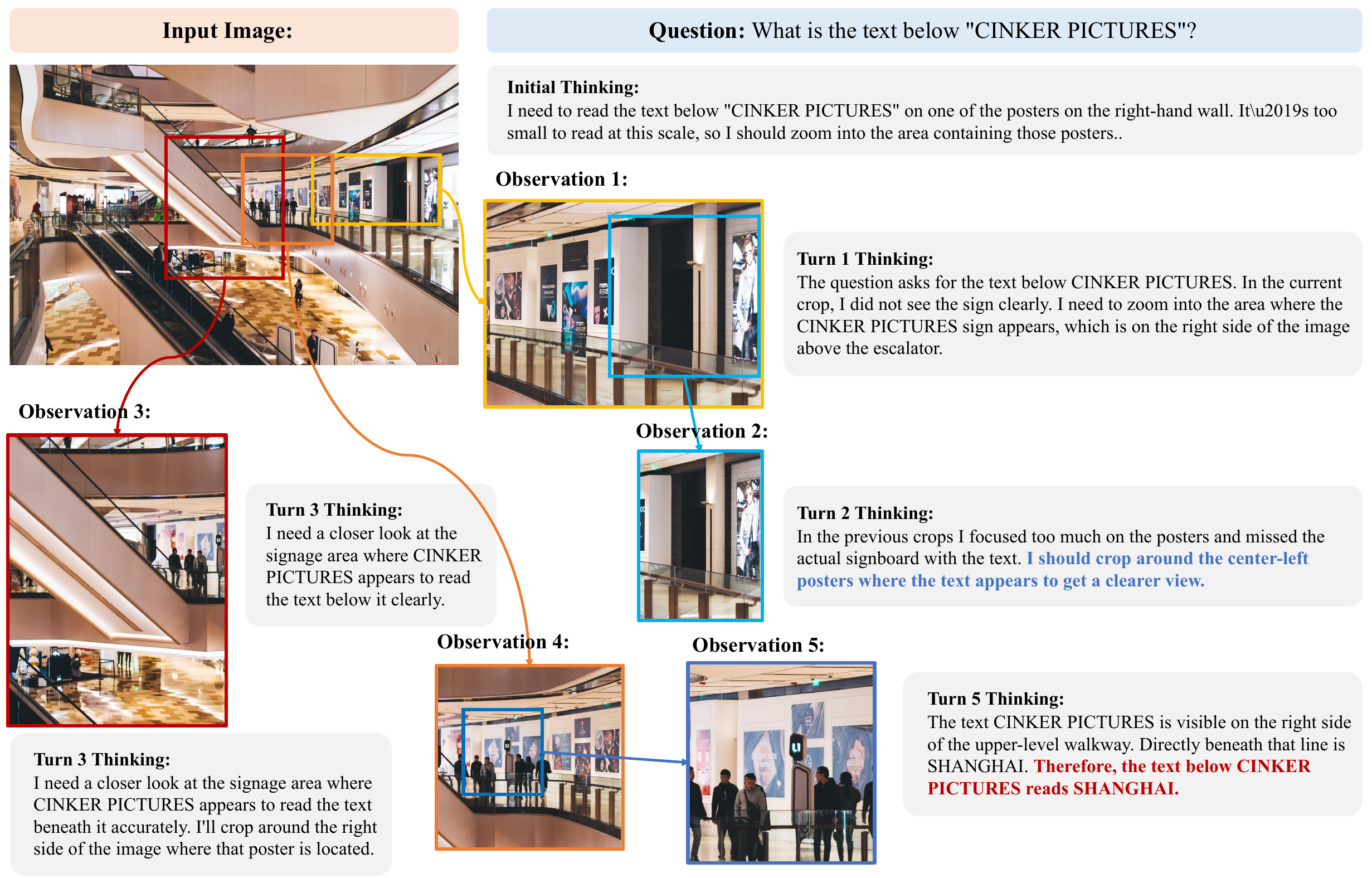}}
\caption{Case III: illustration of multi-turn reasoning}
\label{fig:case13}
\end{center}
\end{figure}

\section{Future Work}
\label{supp:fut}
There are several promising future directions for further advancing the “think with images” capability in visual reasoning. 
\begin{itemize}
    \item \textbf{Overthinking:} The overthinking issue is a well-known limitation of tool-augmented reasoning methods. Specifically, the model tends to invoke tools excessively even when it has already obtained sufficient information to answer correctly, which leads to unnecessarily long reasoning chains when thinking with images in its MCoT. 
    \item \textbf{Multi-tool coordination:} Beyond workflow-predefined approaches, most existing ``thinking with images'' systems support only a single crop-based tool. However, an ideal capability is for the model to autonomously invoke a diverse set of visual tools, such as drawing auxiliary lines or planning maze paths as demonstrated by o3. Achieving such multi-tool coordination is highly challenging, since introducing multiple tools substantially increases the complexity of the learning problem and the difficulty of discovering effective tool-usage strategies. 
    \item \textbf{Visual Hallucination:} When analyzing the failure cases of DRIM, we observe that even after performing multi-scale exploration and correctly localizing the target region, the model may still produce incorrect answers due to limitations in visual perception. A stable visual reasoning CoT should be expected to output the correct answer whenever the target region has been accurately identified in the image. 
\end{itemize}

\end{document}